\documentclass{article}

\usepackage{microtype}
\usepackage{graphicx}
\usepackage{subcaption}
\usepackage{booktabs} 

\usepackage{hyperref}

\usepackage[preprint]{icml2026}

\usepackage{amsmath}
\usepackage{amssymb}
\usepackage{mathtools}
\usepackage{amsthm}

\usepackage[capitalize,noabbrev]{cleveref}
\usepackage{graphicx}
\usepackage{amsmath,amssymb}
\usepackage{caption}
\usepackage{float}    

\theoremstyle{plain}

\theoremstyle{definition}

\theoremstyle{remark}

\usepackage[textsize=tiny]{todonotes}

\icmltitlerunning{KVSlimmer: Theoretical Insights and Practical Optimizations for Asymmetric KV Merging}

\begin{document}

\twocolumn[
  \icmltitle{KVSlimmer: Theoretical Insights and Practical Optimizations for Asymmetric KV Merging}




\icmlsetsymbol{equal}{*}

\begin{icmlauthorlist}
  \icmlauthor{Lianjun Liu}{hu-ice}
  \icmlauthor{Hongli An}{hu-css}
  \icmlauthor{Weiqi Yan}{xmu-mac}
  \icmlauthor{Xin Du}{hu-cst}
  \icmlauthor{Shengchuan Zhang}{xmu-mac}
  \icmlauthor{Huazhong Liu}{hu-cst}
  \icmlauthor{Yunshan Zhong}{hu-cst,equal} 
\end{icmlauthorlist}

\icmlaffiliation{hu-ice}{School of Information and Communication Engineering, Hainan University, Haikou, China}
\icmlaffiliation{hu-css}{School of Cyberspace Security, Hainan University, Haikou, China}
\icmlaffiliation{xmu-mac}{MAC Lab, Department of Artificial Intelligence, School of Informatics, Xiamen University, Xiamen, China}
\icmlaffiliation{hu-cst}{School of Computer Science and Technology, Hainan University, Haikou, China}

\icmlcorrespondingauthor{Yunshan Zhong}{yszhong01@gmail.com}

  \icmlkeywords{Machine Learning, Long-context LLMs, KV cache, Compression}
  \vskip 0.3in
]



\printAffiliationsAndNotice{}  

\begin{abstract}

The growing computational and memory demands of the Key‑Value (KV) cache significantly limit the ability of Large Language Models (LLMs). While KV merging has emerged as a promising solution, existing methods that rely on empirical observations of KV asymmetry and gradient-based Hessian approximations lack a theoretical foundation and incur suboptimal compression and inference overhead.
To bridge these gaps, we establish a theoretical framework that characterizes this asymmetry through the spectral energy distribution of projection weights, demonstrating that concentrated spectra in Query/Key weights induce feature homogeneity, whereas dispersed spectra in Value weights preserve heterogeneity.
Then, we introduce KVSlimmer, an efficient algorithm that captures exact Hessian information through a mathematically exact formulation, and derives a closed-form solution utilizing only forward-pass variables, resulting in a gradient-free approach that is both memory- and time-efficient.
Extensive experiments across various models and benchmarks demonstrate that KVSlimmer consistently outperforms SOTA methods. For instance, on Llama3.1-8B-Instruct, it improves the LongBench average score by 0.92 while reducing memory costs and latency by 29\% and 28\%, respectively.Code is available at \url{https://github.com/lianjunl13-sudo/KVSlimmer}.
\end{abstract}

\section{Introduction}

Large Language Models (LLMs) are increasingly tasked with processing long contexts for applications such as multi-step tool usage, retrieval-augmented generation based on multi-document corpora, chain-of-thought style reasoning, coding agents, and so on~\cite{ijcai.2024/917,liu2025surveytransformercontextextension,huang2024advancingtransformerarchitecturelongcontext,liu2025spakelongcontextlargelanguage}.
However, as the context length extends, the quadratic computational growth of the attention mechanism and the linear expansion of the Key-Value (KV) cache storage~\cite{NEURIPS2022_67d57c32,abs-2209-04881} create a severe memory bottleneck, hindering the practical deployment of LLMs for ultra-long sequences.

\begin{figure}
    \centering
    \includegraphics[width=1\linewidth]{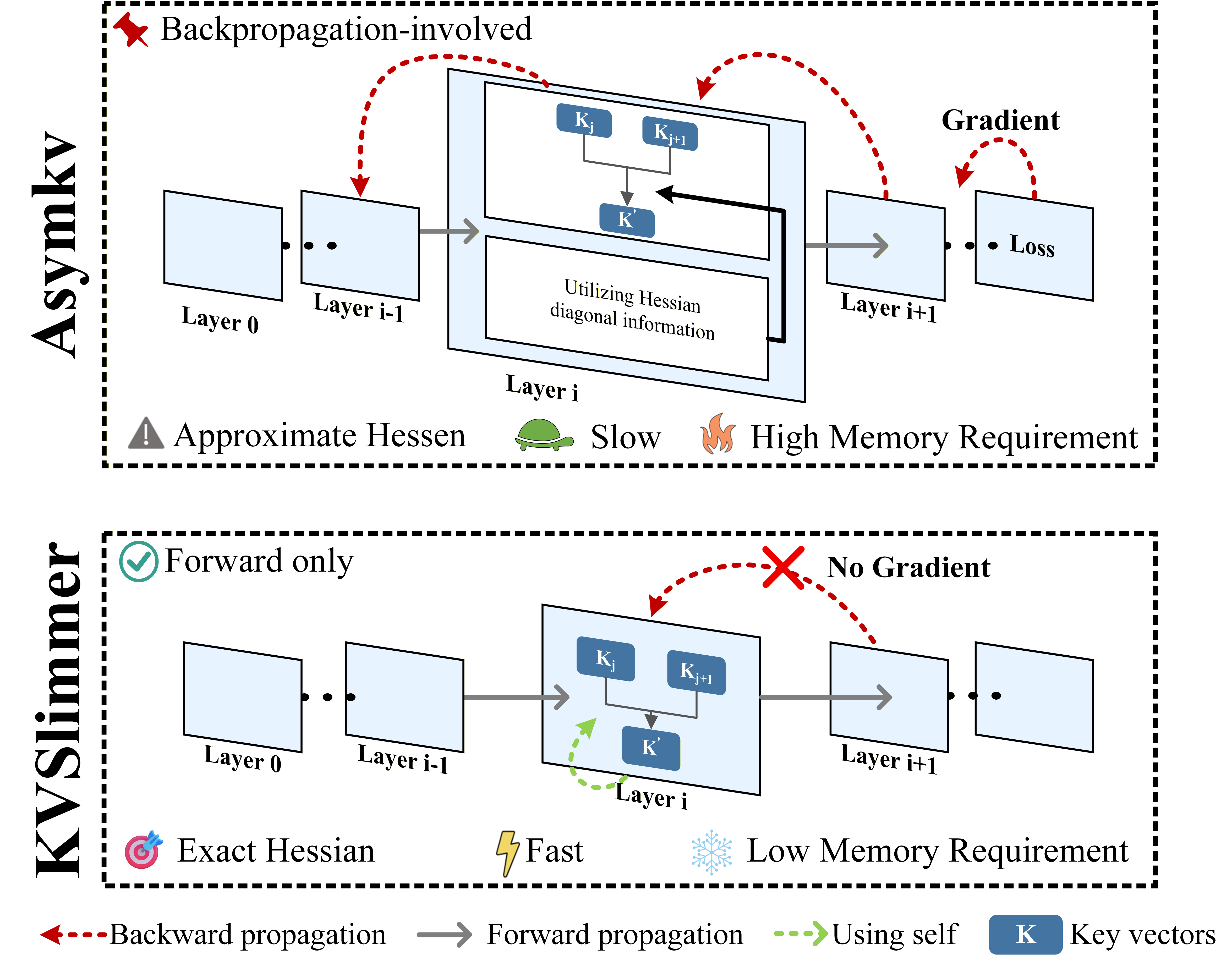}
    \caption{Comparison between AsymKV and KVSlimmer for KV cache merging.}
    \label{fig:framework}
\vskip -10pt
\end{figure}

To mitigate this, KV cache compression has emerged as a pivotal solution. Existing approaches primarily fall into two categories: eviction and merging. Eviction methods~\cite{h2o,Scissorhands,ge2024model,attention_sink} prune tokens deemed less important, but risk discarding information critical for future predictions. Merging methods~\cite{zhang2024cam,liu2024minicache,wang2025model,wan2025textdtexto}, which combine multiple tokens into condensed representations, offer a more information-preserving alternative.

While conventional KV merging methods often apply identical operations to Keys and Values, the recent AsymKV~\cite{asymkv} empirically revealed a critical asymmetry: adjacent Keys exhibit high homogeneity, whereas adjacent Values remain markedly heterogeneous.
Building on this insight, AsymKV employs an approximate Hessian that relies on gradient backpropagation for adjacent Keys merging. Nevertheless, it leaves several critical avenues for further investigation: (1) the lack of a theoretical explanation for this asymmetry, (2) an incomplete second-order Hessian approximation that neglects off-diagonal Key couplings, and (3) a practical dependence on backpropagation, incurring inference overhead.

%
%

To bridge these gaps, we first establish a unified spectral analysis framework to uncover the origins of QKV (dis)similarity. Our analysis demonstrates that the homogeneity is fundamentally dictated by the spectral energy distribution of the projection weight, where the concentrated spectral energy in Q/K projections induces homogeneity, whereas the dispersed energy in V projections induces heterogeneity.
Then, we propose KVSlimmer
a theoretically grounded and computationally efficient framework for asymmetric KV cache merging. As shown in Fig.~\ref{fig:framework}, KVSlimmer derives the exact Hessian to explicitly capture the off-diagonal coupling between adjacent Keys and, more importantly, eliminates the need for backpropagation by deriving a closed-form solution that relies solely on forward-pass variables. This results in a gradient-free, memory- and time-efficient merging algorithm that is both mathematically precise and practically lightweight. Among them, Fig.~\ref{fig:framework} illustrates the differences between KVSlimmer and AsymKV.
Extensive experiments across multiple models and benchmarks demonstrate that KVSlimmer consistently outperforms existing SOTA methods. For instance, when applying KVSlimmer on Llama3.1-8B-Instruct with a chunk\_size of 512, it improves the LongBench average score by 0.92 while still reducing memory costs and latency by 29\% and 28\%, respectively.

\section{Related Work}

\subsection{Long-context Segmentation and Sliding}

As a popular KV compression paradigm, long-context segmentation methods partition the context into multiple segments and retain long-range information dependencies by cross-segment recurrence~\cite{dai-etal-2019-transformer} or by explicit memory~\cite{Rae2020Compressive}. 
RMT~\cite{10.5555/3600270.3601075} strengthens cross-chunk integration by adopting memory tokens to deliver long-range information.
Another paradigm is long-context sliding~\cite{ZaheerGDAAOPRWY20,ainslie-etal-2020-etc}, which controls KV overhead by adopting local-window attention or sparse attention within each chunk. Longformer~\cite{abs-2004-05150} combines sliding-window attention with a small set of global tokens, forming an efficient baseline for long-document modeling. 
%
%
%
Recent studies~\cite{zhu-etal-2024-coca,wu-etal-2025-tokenselect} improve the practicality and stability of segmentation and sliding at inference time.
StreamingLLM~\cite{attention_sink} stabilizes sliding-window inference via a fixed set of initial KV tokens, known as attention sinks. 
DCA~\cite{abs-2402-17463} decomposes attention into intra-block and inter-block modules to achieve training-free long-context extension. 
InfLLM~\cite{10.5555/3737916.3741717} extrapolates to long contexts by leveraging efficient context memory and retrieval mechanisms. 
In addition, CoCA~\cite{zhu-etal-2024-coca} improves boundary behaviors in long-context extrapolation by addressing the coupling between positional encoding and attention. TokenSelect~\cite{wu-etal-2025-tokenselect} selectively involves a few critical KV in attention calculation to reduce inference cost.  
%

\subsection{KV cache eviction.}
KV eviction methods reduce the computational load and memory footprint during the decoding process initially~\cite{gu2025obcacheoptimalbrainkv,kim2025epicacheepisodickvcache,chittyvenkata2025pagedevictionstructuredblockwisekv,wang2025lookaheadqcacheachievingconsistent,liu2023scissorhands}. 
Subsequently, the focus has gradually shifted towards importance-based KV eviction strategies. 
H$_2$O~\cite{h2o} identifies and prioritizes the KV with high contribution evaluated by accumulated attention scores. Subsequently, Scissorhands~\cite{liu2023scissorhands} further introduces the ``importance persistence'' assumption, retaining KV tokens in a probabilistic manner under a fixed cache budget. 
Another series of methods moves the eviction decision forward to the prefill phase. For example, SnapKV~\cite{10.5555/3737916.3738638} and FastGen~\cite{ge2023model} estimate the long-term contribution of prompt tokens through observation windows and attention head differences, respectively. 
To maximize the utilization efficiency under limited budgets, PyramidKV~\cite{cai2025pyramidkv} and HeadKV~\cite{fu2025not} introduce layer- and head-level budget allocation strategies.
However, as KV eviction methods rely on historical importance estimates to permanently discard tokens, they fail to account for the temporal dynamic nature of token importance, where tokens historically deemed insignificant can become pivotal for future predictions~\cite{gu2025ahakvadaptiveholisticattentiondriven,feng2025evicpressjointkvcachecompression,feng2025tamingfragilitykvcache}.

\subsection{KV Cache Merging.}
KV merging methods aggregate multiple historical KV tokens into fewer representations, reducing the KV overhead while still retaining more contextual information for future predictions~\cite{ancucki2025inferencetime,brandon2024reducing,yuan2025weightedkvattentionscoresweighted,li2025emsadaptiveevictthenmergestrategy,wang2024modeltellsmergeadaptive}.
CaM~\cite{zhang2024cam} aggregates KV tokens by employing the ratio of attention scores as merging weights, ensuring that historical importance is preserved within the compressed representation.
DMC~\cite{10.5555/3692070.3693589} learns to decide during generation whether to write new entries to the cache or merge them with existing cache entries, adaptively controlling compression intensity across different layers and heads.
For long-context understanding evaluation, KVMerger~\cite{wang2025model} models the identification of merge token sets as a constrained clustering problem and employs a kernel-weighted merging strategy. 
D$_2$O~\cite{wan2025textdtexto} dynamically optimizes KV cache size at both the layer- and token-level. 
%
Most existing KV merging methods assume they are functionally equivalent and apply a unified merging strategy to both of them~\cite{li2025flowmmcrossmodalinformationflow,tian2025keepkvachievingperiodiclossless,chang2025xkvcrosslayersvdkvcache,liu2025zsmergezeroshotkvcache}. But the recent AsymKV~\cite{asymkv}, going beyond a unified strategy, empirically uncovers a structural asymmetry within the adjacent KV cache, where Keys exhibit high homogeneity while Values remain heterogeneous. Based on this finding, AsymKV mathematically proposes a Hessian-based merging strategy that fuses redundant Keys and lossless cardinality normalization for Values merging.

\begin{figure*}[t]
  \begin{center}
    \makebox[\textwidth][c]{%
      \includegraphics[
        width=0.9\textwidth,
      ]{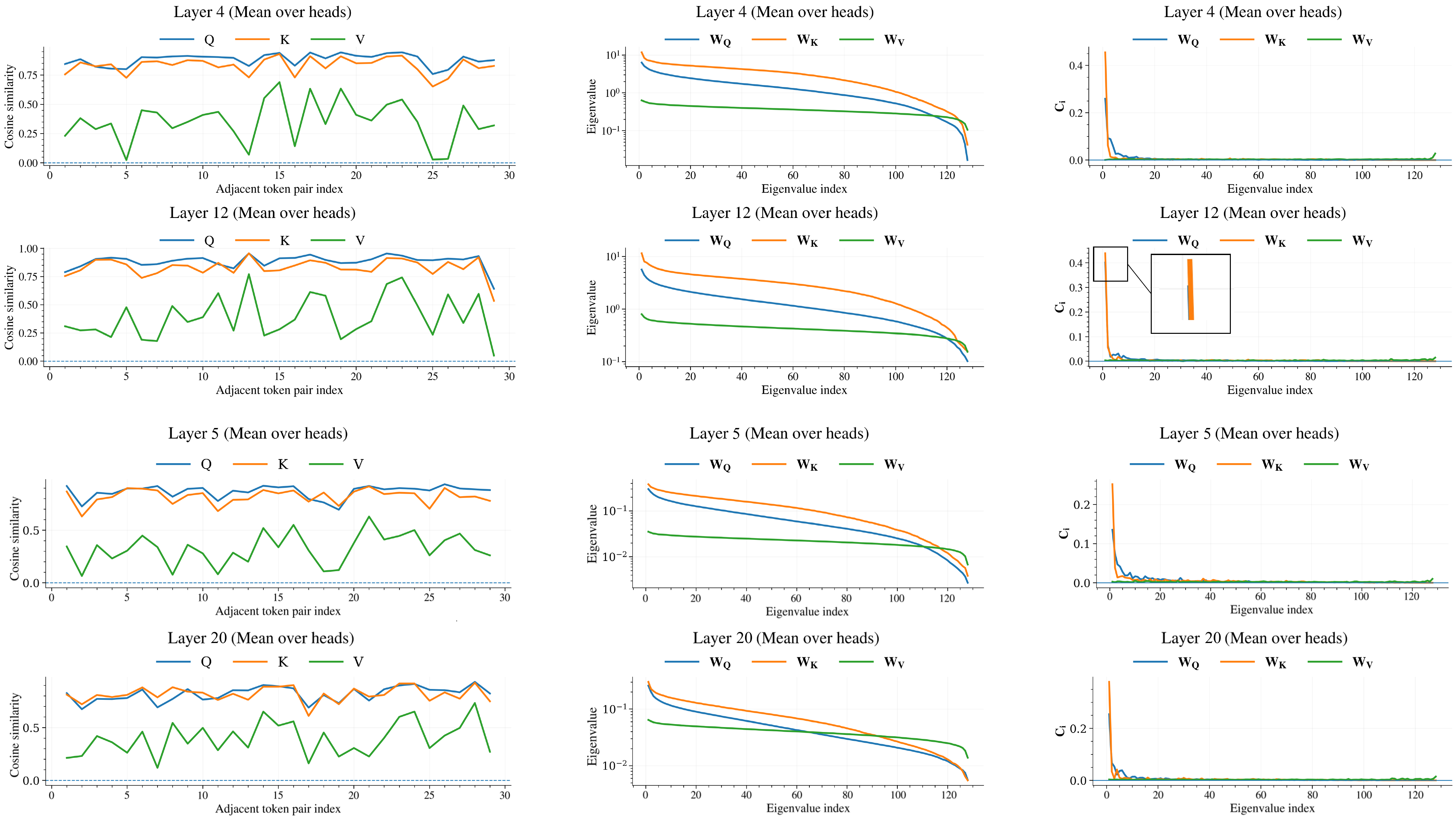}
    }
    \caption{
Layer-wise QKV similarity and spectral analysis.
\textbf{Left column:} Mean adjacent-token cosine similarity for Query (Q), Key (K), and Value (V), averaged over attention heads.
\textbf{Middle column:} Eigenvalue distributions of the projection matrices $\mathbf{W}_Q$, $\mathbf{W}_K$, and $\mathbf{W}_V$,
sorted in descending order.
\textbf{Right column:} Mode-wise contribution coefficients $c_i$ (Eq.~\ref{eq:mode_contribution}), plotted according to the eigenvalue index.
The first two rows show results from \textit{Llama-3.1-8B-Instruct},
while the last two rows show results from \textit{Mistral-7B-Instruct-v0.3}.
}
\label{fig:QKVsimilarity}
\end{center}
\vskip -10pt
\end{figure*}

\section{Root of KV Asymmetry}

\subsection{Preliminary}

In this subsection, we briefly review AsymKV~\cite{asymkv}, a recent KV merging method that serves as a starting point for our KVSlimmer.
AsymKV is motivated by the empirical observation that adjacent Keys typically exhibit high homogeneity, while adjacent Values are markedly heterogeneous. Consequently, it proposes a Hessian-based non-uniform merging strategy specifically for adjacent Keys.
Given a sequence of Keys $\mathbf{K}=[\mathbf{k}_1, \dots, \mathbf{k}_n]$, AsymKV aims to merge adjacent $(\mathbf{k}_m, \mathbf{k}_{m+1})$ into a single Key $\mathbf{k}^*$. To determine the optimal $\mathbf{k}^*$, it minimizes the following loss:
\begin{equation}
\mathbf{k}^* = \arg\min_{\mathbf{k}} \mathcal{L}(\mathbf{k}, \mathbf{k}),
\end{equation}
where $\mathcal{L}(\mathbf{k}, \mathbf{k})$ denotes the loss when the $(\mathbf{k}_m, \mathbf{k}_{m+1})$ is replaced by $(\mathbf{k}, \mathbf{k})$.
To solve this optimization problem, a second-order Taylor expansion is applied at the original points $(\mathbf{k}_m, \mathbf{k}_{m+1})$. Let $\mathbf{h}^{a,b}\in \mathbb{R}^{d \times d}$ be the Hessian matrix between $(\mathbf{k}_a, \mathbf{k}_{b})$. By using a modified Newton approach that maintains numerical stability, the solution for the optimal $\mathbf{k}^*$ is derived:
\begin{equation} 
\begin{aligned}
\mathbf{k}^* 
&= (\mathbf{h}^{m,m}+2\mathbf{h}^{m,m+1}+\mathbf{h}^{m+1,m+1})^{-1} \\
&\quad\; \Big[
    \mathbf{h}^{m,m}\mathbf{k}_m
    + \mathbf{h}^{m,m+1}(\mathbf{k}_m+\mathbf{k}_{m+1}) \\
&\quad\; + \mathbf{h}^{m+1,m+1}\mathbf{k}_{m+1} \Big].
\end{aligned}
\label{eq:optimal_k_cn}
\end{equation}
In practice, the off-diagonal Hessian block $\mathbf{h}^{m,m+1}$ is ignored, and the diagonal elements are approximated using the Fisher Information Matrix (i.e., $\mathbf{h}^{ii} \approx (\nabla_i \mathcal{L})^2$). Under these simplifications, the optimal Key is constructed as a weighted aggregation:
\begin{equation}\mathbf{k}^* \approx (\mathbf{h}^{m,m} + \mathbf{h}^{m+1,m+1})^{-1} (\mathbf{h}^{m,m}\mathbf{k}_m + \mathbf{h}^{m+1,m+1}\mathbf{k}_{m+1}).\label{eq:simplified_k}
\end{equation}
For a merged adjacent pair $(m,m{+}1)$, the Value is combined by simple addition:
\begin{equation}
\mathbf v^* = \mathbf v_m + \mathbf v_{m+1}.
\end{equation}
To optimize throughput, a chunk-based compression is implemented. Upon reaching the \texttt{budget}, AsymKV merges \texttt{chunk\_size} adjacent pairs in parallel per new generated chunk, restoring the sequence length from \texttt{budget + chunk\_size} back to the budget in a single step.

While current methods have made significant strides, they leave three key challenges unaddressed: (1) the theoretical explanation of KV asymmetry remains underexplored; (2) existing Hessian approximations overlook the off-diagonal couplings between Keys; and (3) a practical reliance on backpropagation persists, leading to non-negligible inference overhead. In this paper, we aim to address these challenges.

\subsection{Theoretical Analysis of QKV Homogeneity and Heterogeneity}
\label{projection}

In this subsection, we establish a unified framework for QKV (dis)similarity by linking the spectral energy distribution of their projection weight matrices to their functional outcomes. Specifically, we show that projections that concentrate spectral energy induce adjacent homogeneity, whereas those with dispersed energy preserve heterogeneity.

Let $\mathbf{X} = \{\mathbf{x}_t\}_{t=1}^{T}$ denote the hidden-state sequence. For a specific attention head, the linear projection for the Query, Key, or Value can be generically represented as $\mathbf{y}_t = \mathbf{x}_t \mathbf{W}$, where $\mathbf{y} \in \{\mathbf{Q}, \mathbf{K}, \mathbf{V}\}, \mathbf{W} \in \{\mathbf{W}_\mathbf{Q}, \mathbf{W}_\mathbf{K}, \mathbf{W}_\mathbf{V}\}$.
Then, the cosine similarity between adjacent projected tokens $(\mathbf{x}_t, \mathbf{x}_{t+1})$ is defined as:
\begin{equation} 
\label{eq:cosine_raw} 
\cos(\mathbf{y}_t, \mathbf{y}_{t+1}) = \frac{(\mathbf{x}_t \mathbf{W})(\mathbf{x}_{t+1} \mathbf{W})^\top}{\|\mathbf{x}_t \mathbf{W}\|_2 \|\mathbf{x}_{t+1} \mathbf{W}\|_2}. 
\end{equation}

By defining the induced metric $\mathbf{M} \triangleq \mathbf{W} \mathbf{W}^\top$, we can reparameterize Eq.~\ref{eq:cosine_raw} within the input space:
\begin{equation}
\label{eq:cosine_M}
\cos(\mathbf{y}_t, \mathbf{y}_{t+1}) = \frac{\mathbf{x}_t \mathbf{M} \mathbf{x}_{t+1}^\top}{\sqrt{\mathbf{x}_t \mathbf{M} \mathbf{x}_t^\top} \sqrt{\mathbf{x}_{t+1} \mathbf{M} \mathbf{x}_{t+1}^\top}}.
\end{equation}

To elucidate the influence of the projection's spectral properties, we apply Singular Value Decomposition (SVD) to the weight matrix $\mathbf{W} = \mathbf{U} \mathbf{\Sigma} \mathbf{V}^\top$, where $\mathbf{\Sigma} = \mathrm{diag}(\sigma_1, \dots, \sigma_d)$. The induced metric $\mathbf{M}$ thus admits the spectral decomposition $\mathbf{M} = \mathbf{U} \mathbf{\Sigma}^2 \mathbf{U}^\top = \sum_{i=1}^{d} \lambda_i \mathbf{u}_i \mathbf{u}_i^\top$, with $\lambda_i = \sigma_i^2$.

By projecting the input $\mathbf{x}_t$ onto the modal coordinates spanned by the left singular vectors, denoted as $p_{t,i} \triangleq \mathbf{x}_t \mathbf{u}_i$, the bilinear form can be expanded as $\mathbf{x}_t \mathbf{M} \mathbf{x}_{t+1}^\top = \sum_{i=1}^d \lambda_i p_{t,i} p_{t+1,i}$. Consequently, the cosine similarity admits an exact decomposition into the contributions of individual spectral modes:
\begin{equation}
\cos(\mathbf{y}_t, \mathbf{y}_{t+1}) = \frac{\sum_{i=1}^{d} \lambda_i p_{t,i} p_{t+1,i}}{\sqrt{\sum_{i=1}^{d} \lambda_i p_{t,i}^2} \sqrt{\sum_{i=1}^{d} \lambda_i p_{t+1,i}^2}}.
\label{eq:cosine_spectral}
\end{equation}

Furthermore, we define the relative contribution of the $i$-th spectral mode as:
\begin{equation}
\label{eq:mode_contribution}
c_i(\mathbf{y}_t, \mathbf{y}_{t+1}) \triangleq \frac{\lambda_i p_{t,i} p_{t+1,i}}{\sqrt{\sum_{j=1}^d \lambda_j p_{t,j}^2} \sqrt{\sum_{j=1}^d \lambda_j p_{t+1,j}^2}},
\end{equation}
which simplifies the similarity to a cumulative summation: $\cos(\mathbf{y}_t, \mathbf{y}_{t+1}) = \sum_{i=1}^{d} c_i(\mathbf{y}_t, \mathbf{y}_{t+1})$. 

Since the spectral modes are ordered by decreasing eigenvalues $\lambda_i$, this formulation directly reveals that adjacent similarity is predominantly shaped by high-energy components, provided that the weight matrix exhibits a sharp spectral decay. As illustrated in Fig.~\ref{fig:QKVsimilarity}\footnote{ More illustrations across different models are provided in Appendix.~\ref{A}.}, we empirically observe that $\mathbf{W}_\mathbf{Q}$ and $\mathbf{W}_\mathbf{K}$ projection weights possess a highly concentrated energy spectrum, inducing homogeneity by forcing adjacent embeddings into a shared subspace, while $\mathbf{W}_\mathbf{V}$ projection possesses a relatively dispersed energy spectrum, resulting the heterogeneity.

This theoretical insight reveals an intriguing phenomenon within the attention mechanism. The Query and Key projections are inherently geared toward alignment. Their concentrated spectra effectively filter out high-frequency noise and project tokens into a shared semantic subspace, thereby inducing the stable similarity necessary for robust matching.
In contrast, the Value projection is primarily responsible for information transmission. Its dispersed spectrum preserves the intrinsic heterogeneity, ensuring that the aggregated context remains expressive and information-rich rather than collapsing into a homogenized representation.

\section{KVSlimmer}

\subsection{Exact Hessian Derivation for Key-Key Coupling}


In this subsection, we formulate and derive the exact Hessian, explicitly capturing both the diagonal and off-diagonal coupling between adjacent keys, which prior art has overlooked.

Let $\mathbf{q}$ be the Query, $\mathbf{k}_i, \mathbf{v}_i$ be the Key and Value of the $i$-th token, respectively. The attention logit $e_i$, score $\alpha_i$, and output $o$ are defined as:
\begin{equation}
e_i \,=\, \frac{\mathbf q\,\mathbf k_i^\top}{\sqrt{d_k}},
\alpha_i \,=\, \frac{\exp e_i}{\sum_{t=1}^n \exp e_t}, 
\mathbf{o}        \,=\, \sum_{t=1}^n \alpha_t \mathbf{v}_t,
\label{eq:attn_def_vec}
\end{equation}
where $n$ is the sequence length. We denote the loss by $\mathcal{L}$ and let $\mathbf{E} = \partial \mathcal{L}/\partial \mathbf{o}$ be the gradient of the loss with respect to the attention output.

We first derive the gradient $\mathbf{g}_i = \nabla_{\mathbf{k}_i} \mathcal{L}$. The Jacobian of the softmax is $\partial \alpha_j/\partial e_i = \alpha_j(\delta_{ji} - \alpha_i)$. Since $e_i$ depends only on $\mathbf{k}_i$, we have:
\begin{equation}
\frac{\partial e_i}{\partial \mathbf{k}_i} = \frac{\mathbf{q}}{\sqrt{d_k}}, \qquad
\frac{\partial e_m}{\partial \mathbf{k}_i} = \mathbf{0} \ \text{for}\ m \neq i.
\end{equation}

Using the chain rule, the gradient of the output with respect to a key vector is:
\begin{equation}
\frac{\partial \mathbf{o}}{\partial \mathbf{k}_i}
= \sum_{j=1}^n \mathbf{v}_j \frac{\partial \alpha_j}{\partial \mathbf{k}_i}
= \frac{1}{\sqrt{d_k}} \alpha_i (\mathbf{v}_i - \mathbf{o}) \mathbf{q}^\top.
\end{equation}

Therefore, the loss gradient with respect to $\mathbf{k}_i$ is:
\begin{equation}
\mathbf{g}_i = \nabla_{\mathbf{k}_i} \mathcal{L}
= \frac{\partial \mathcal{L}}{\partial \mathbf{o}} \frac{\partial \mathbf{o}}{\partial \mathbf{k}_i}
= \frac{1}{\sqrt{d_k}} \alpha_i \left[ \mathbf{E}^\top (\mathbf{v}_i - \mathbf{o}) \right] \mathbf{q}.
\label{eq:key_gradient}
\end{equation}

We now compute the Hessian block $\mathbf{h}^{ij} = \partial \mathbf{g}_i / \partial \mathbf{k}_j^\top$, which captures the second-order interaction between the $i$-th and $j$-th Keys.
Define $\mathbf{s_i} = \alpha_i (\mathbf{v}_i - \mathbf{o})$, so that $\mathbf{g}_i = \frac{1}{\sqrt{d_k}} (\mathbf{E}^\top \mathbf{s_i}) \mathbf{q}$, then:
\begin{equation}
\mathbf{h}^{ij} = \frac{1}{\sqrt{d_k}} \mathbf{q} \frac{\partial [\mathbf{E}^\top \mathbf{s_i}]}{\partial \mathbf{k}_j^\top}.
\label{eq:hessian_intermediate}
\end{equation}
Since $e_j = \mathbf{q}^\top \mathbf{k}_j / \sqrt{d_k}$, we have $\partial e_j / \partial \mathbf{k}_j^\top = \mathbf{q}^\top / \sqrt{d_k}$. Applying the chain rule:
\begin{equation}
\frac{\partial \mathbf{s_i}}{\partial \mathbf{k}_j^\top}
= \frac{\partial \mathbf{s_i}}{\partial e_j} \frac{\partial e_j}{\partial \mathbf{k}_j^\top}
= \frac{1}{\sqrt{d_k}} \frac{\partial \mathbf{s_i}}{\partial e_j} \mathbf{q}^\top.
\end{equation}
Substituting into Eq.~\ref{eq:hessian_intermediate} and using the linearity of $\mathbf{E}^\top$, we obtain:
\begin{equation}
\mathbf{h}^{ij} = \frac{1}{d_k} \left[ \mathbf{E}^\top \frac{\partial \mathbf{s_i}}{\partial e_j} \right] \mathbf{q} \mathbf{q}^\top.
\label{eq:hessian_rank1}
\end{equation}

Thus, each Hessian block $\mathbf{h}^{ij}$ is a rank-one matrix $\mathbf{q}\mathbf{q}^\top$ scaled by a scalar coefficient.
%
To compute $\partial \mathbf{s}_i / \partial e_j$, expand $\mathbf{s_i} = \alpha_i \mathbf{v}_i - \alpha_i \mathbf{o}$:
\begin{equation}
\frac{\partial \mathbf{s_i}}{\partial e_j}
= \frac{\partial \alpha_i}{\partial e_j} \mathbf{v}_i -\frac{\partial \alpha_i} {\partial e_j} \mathbf{o} - \alpha_i \frac{\partial \mathbf{o}}{\partial e_j}.
\end{equation}
From the definition of $\mathbf{o}$ and the softmax Jacobian, we have:
\begin{equation}
\frac{\partial \mathbf{o}}{\partial e_j} = \sum_{t=1}^n \mathbf{v}_t \frac{\partial \alpha_t}{\partial e_j}
= \alpha_j (\mathbf{v}_j - \mathbf{o}).
\end{equation}
Combining these results yields two distinct cases:

(1) Diagonal case ($j=i$), which captures the \textit{self-sensitivity} of the $i$-th Key:
\begin{equation}\frac{\partial \mathbf{s}_i}{\partial e_i} = \alpha_i (1 - 2\alpha_i)(\mathbf{v}_i - \mathbf{o}).
\end{equation}
(2) Off-diagonal case ($j \neq i$), which captures the \textit{coupling information} between $i$-th and $j$-th Keys:
\begin{equation}\frac{\partial \mathbf{s}_i}{\partial e_j} = -\alpha_i \alpha_j (\mathbf{v}_i + \mathbf{v}_j - 2\mathbf{o}).\end{equation}

Thus, for any two adjacent Keys $(\mathbf{k}_m, \mathbf{k}_{m+1})$ targeted for merging, the precise Hessians are recovered as:
%
\begin{align}
\mathbf{h}^{mm} &= \frac{1}{d_k} \left[ \mathbf{E}^\top \alpha_m (1-2\alpha_m)(\mathbf{v}_m - \mathbf{o}) \right] \mathbf{q} \mathbf{q}^\top, 
\label{eq:h11_mat} \\
\mathbf{h}^{m+1,m+1} &= \frac{1}{d_k} \left[ \mathbf{E}^\top \alpha_{m+1} (1-2\alpha_{m+1})(\mathbf{v}_{m+1} - \mathbf{o}) \right] \mathbf{q} \mathbf{q}^\top, \label{eq:h12_mat} \\
\mathbf{h}^{m,m+1} &= -\frac{1}{d_k} \left[ \mathbf{E}^\top \alpha_m \alpha_{m+1} (\mathbf{v}_m + \mathbf{v}_{m+1} - 2\mathbf{o}) \right] \mathbf{q} \mathbf{q}^\top,
\label{eq:h22_mat}
\end{align}
with $\mathbf h^{m+1,m}=\mathbf h^{m,m+1}$.

\begin{figure}[t]
    \centering
    \begin{subfigure}{0.32\linewidth}
        \centering
        \includegraphics[width=\linewidth]{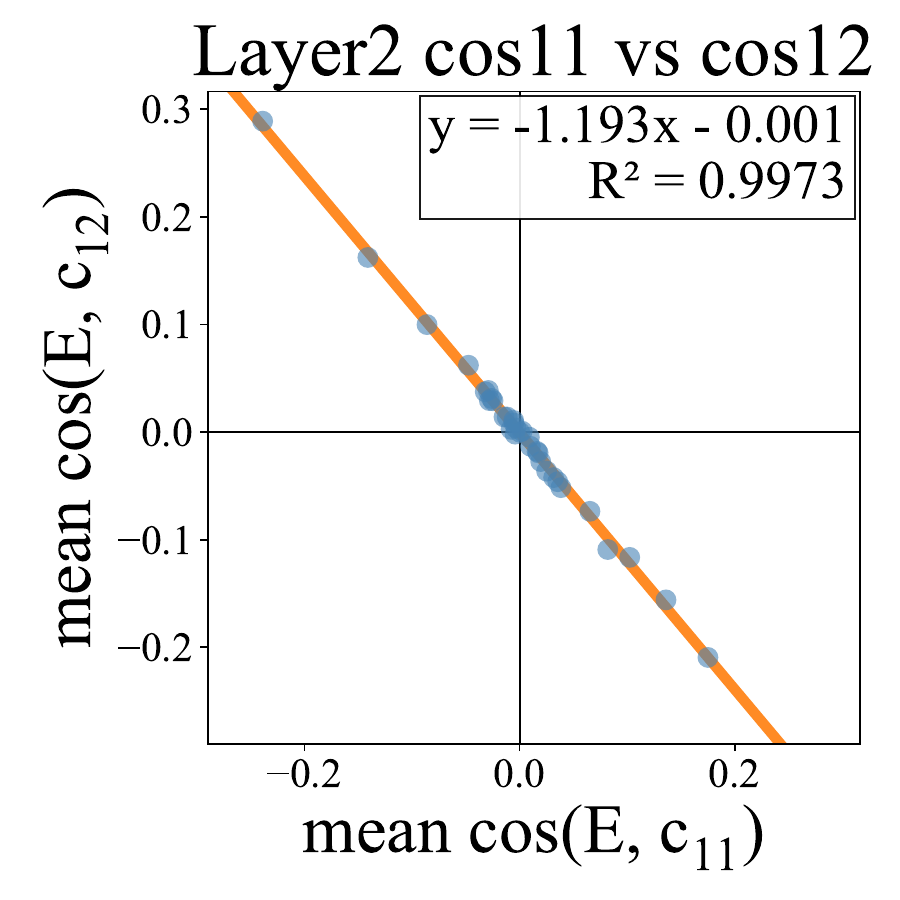}
        \caption{$\cos(\mathbf{E},\mathbf{c}_{11})$ vs. $\cos(\mathbf{E},\mathbf{c}_{12})$}
    \end{subfigure}\hfill
    \begin{subfigure}{0.32\linewidth}
        \centering
        \includegraphics[width=\linewidth]{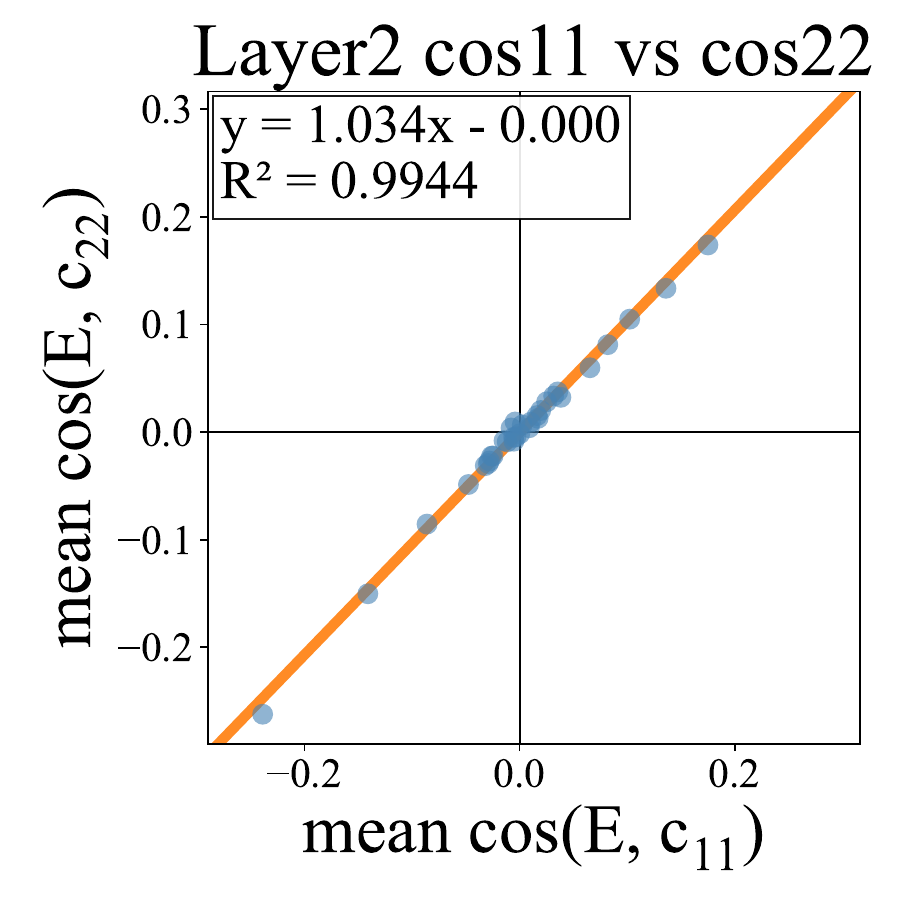}
        \caption{$\cos(\mathbf{E},\mathbf{c}_{11})$ vs. $\cos(\mathbf{E},\mathbf{c}_{22})$}
    \end{subfigure}\hfill
    \begin{subfigure}{0.32\linewidth}
        \centering
        \includegraphics[width=\linewidth]{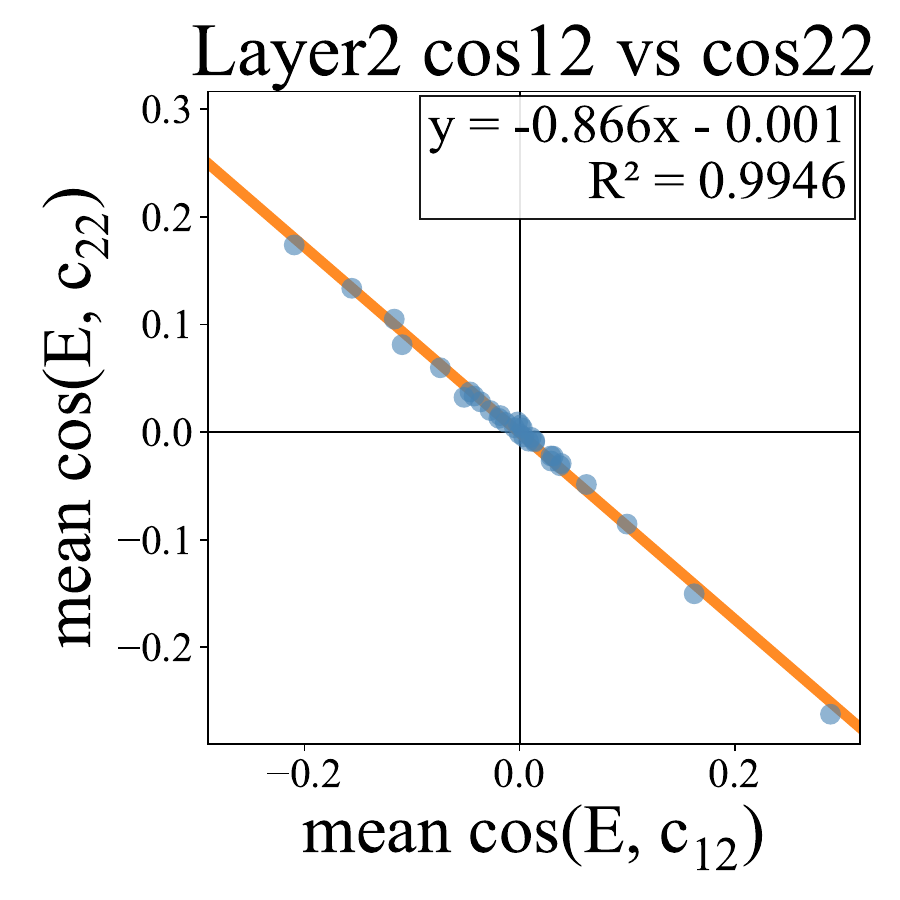}
        \caption{$\cos(\mathbf{E},\mathbf{c}_{12})$ vs. $\cos(\mathbf{E},\mathbf{c}_{22})$}
    \end{subfigure}
    \caption{Head-level mean alignment relationships at Layer~2 of \textit{Llama-3.1-8B-Instruct} on 2WikiMQA. Each point corresponds to one attention head, positioned by its global mean cosine alignment.}
    \label{fig:layer2_head_alignment}
\vskip -10pt
\end{figure}

\subsection{Computation simplification}

The exact Hessian blocks derived in Eqs.~\ref{eq:h11_mat}-\ref{eq:h22_mat} enable precise computation of the optimal merged key in Eq.~\ref{eq:optimal_k_cn}. However, this requires the loss gradient $\mathbf{E}=\partial\mathcal{L}/\partial\mathbf{o}$, necessitating expensive backpropagation. In this subsection, we eliminate gradient dependence, yielding a memory- and compute-efficient solution that preserves Hessian information precisely.

We begin by defining three vectors computed solely from the forward-pass:
\begin{align}
\mathbf{c}_{11} &= \alpha_m(1-2\alpha_m)(\mathbf{v}_m-\mathbf{o}), \label{eq:c11_def}\\
\mathbf{c}_{22} &= \alpha_{m+1}(1-2\alpha_{m+1})(\mathbf{v}_{m+1}-\mathbf{o}), \label{eq:c22_def}\\
\mathbf{c}_{12} &= -\alpha_m\alpha_{m+1}(\mathbf{v}_m+\mathbf{v}_{m+1}-2\mathbf{o}), \label{eq:c12_def}
\end{align}
where $\mathbf{c}_{ij}\in\mathbb{R}^{d_v}$. The Hessian blocks $\mathbf{h}^{ij}$ share a unified, rank-one form governed by scalar sensitivity $g_{ij} \triangleq \mathbf{E}^\top \mathbf{c}_{ij}$:
\begin{equation}
\mathbf{h}^{ij} = \frac{1}{d_k}\,g_{ij}\,\mathbf{q}\mathbf{q}^\top, \quad ij\in\{11,12,22\}.
\label{eq:h_rank1_unified}
\end{equation}

Substituting Eq.~\ref{eq:h_rank1_unified} into Eq.~\ref{eq:optimal_k_cn} yields the linear system $\mathbf{M}\mathbf{k}^* = \mathbf{N}$, where
\begin{equation}
\mathbf M
\triangleq
\mathbf h^{11}+2\mathbf h^{12}+\mathbf h^{22}
=
\frac{1}{d_k}\gamma\,\mathbf q\mathbf q^\top,
\,\,
\gamma\triangleq g_{11}+2g_{12}+g_{22},
\label{eq:H_gamma_rank1_rewrite}
\end{equation}
\begin{equation}
\mathbf N
\triangleq
\mathbf h^{11}\mathbf k_m+\mathbf h^{12}(\mathbf k_m+\mathbf k_{m+1})
+\mathbf h^{22}\mathbf k_{m+1}
=
\frac{1}{d_k}\,\mathbf q\mathbf q^\top\,\mathbf b,
\label{eq:u_rank1_rewrite}
\end{equation}
with $\mathbf{b}\triangleq (g_{11}+g_{12})\mathbf{k}_m+(g_{12}+g_{22})\mathbf{k}_{m+1}$.
Since $\mathbf{q}\mathbf{q}^\top$ is rank-one, a particular solution in $\mathrm{span}\{\mathbf{q}\}$ is obtained via the Moore–Penrose pseudoinverse:
\begin{equation}
\begin{aligned}
\mathbf k^*
&=\mathbf M^{+}\mathbf N
=\Big(\tfrac{1}{d_k}\gamma\,\mathbf q\mathbf q^\top\Big)^{+}
\Big(\tfrac{1}{d_k}\mathbf q\mathbf q^\top\,\mathbf b\Big) \\
&=\frac{1}{\gamma}\,(\mathbf q\mathbf q^\top)^{+}(\mathbf q\mathbf q^\top)\,\mathbf b
=\frac{1}{\gamma}\,\mathbf P_q\,\mathbf b,
\end{aligned}
\label{eq:kstar_proj_form}
\end{equation}
where $\mathbf P_q\triangleq(\mathbf q\mathbf q^\top)^{+}(\mathbf q\mathbf q^\top)$ denotes the orthogonal projection onto $\mathrm{span}\{\mathbf q\}$.

Crucially, the solution $\mathbf{k}^* = \frac{1}{\gamma}\mathbf{P}_q\mathbf{b}$ is confined to the one-dimensional subspace $\mathrm{span}\{\mathbf{q}\}$. Therefore, we can equivalently express it within the original key space $\mathrm{span}\{\mathbf k_m,\mathbf k_{m+1}\}$, which cancels the common factor $\mathbf{q}\mathbf{q}^\top$ and yields a pure weight form:
\begin{equation}
\mathbf k^* = w_m\,\mathbf k_m+w_{m+1}\,\mathbf k_{m+1},
\label{eq:kstar_weight_final_rewrite}
\end{equation}
where $
w_m=\frac{g_{11}+g_{12}}{g_{11}+2g_{12}+g_{22}},
w_{m+1}=\frac{g_{12}+g_{22}}{g_{11}+2g_{12}+g_{22}}. $
Eq.~\ref{eq:kstar_weight_final_rewrite} indicates that the optimal merging weights are determined solely by the relative magnitudes of the scalar sensitivity $g_{ij}$.
The $g_{ij}=\mathbf{E}^\top\mathbf{c}_{ij}$ still depend on the gradient $\mathbf{E}$. Thus, we factor them into norm and angular components:
\begin{equation}
g_{ij}=\mathbf E^\top \mathbf{c}_{ij}
=\|\mathbf E\|_2\,\|\mathbf{c}_{ij}\|_2\,\cos(\mathbf E,\mathbf{c}_{ij}), \, ij\in\{11,12,22\}.
\label{eq:gij_norm_cos}
\end{equation}

In regions where adjacent Keys are homogeneous, our empirical analysis in Fig.~\ref{fig:layer2_head_alignment} reveals a consistent relation:
\begin{equation}
\cos(\mathbf E,\mathbf{c}_{11})
\;\approx\;
\cos(\mathbf E,\mathbf{c}_{22})
\;\approx\;
-\cos(\mathbf E,\mathbf{c}_{12}).
\label{eq:cos_sign_relation_supp1}
\end{equation}

In Appendix.~\ref{B} and ~\ref{C}, we provide more illustrations and a theoretical analysis for this relation, respectively.
%
%
%
Plugging this relation into Eq.~\ref{eq:kstar_weight_final_rewrite}, the common factors cancel exactly and result in:
\begin{equation}
\begin{aligned}
\mathbf{k}^{*}
&=
\Big(
\lVert\mathbf{c}_{11}\rVert_2
-2\lVert\mathbf{c}_{12}\rVert_2
+\lVert\mathbf{c}_{22}\rVert_2
\Big)^{-1}
\\
&\quad\;
\Big[
(\lVert\mathbf{c}_{11}\rVert_2-\lVert\mathbf{c}_{12}\rVert_2)\mathbf k_m
+
(\lVert\mathbf{c}_{22}\rVert_2-\lVert\mathbf{c}_{12}\rVert_2)\mathbf k_{m+1}
\Big].
\end{aligned}
\label{eq:kstar_final}
\end{equation}

Eq.~\ref{eq:kstar_final} provides three key advantages: (1) it eliminates backpropagation and uses only forward-pass variables ($\alpha_i$, $\mathbf{v}_i$, $\mathbf{o}$); (2) it retains essential second-order Key–Key coupling interactions; (3) it involves only norm computations and linear combinations, adding negligible overhead.

\begin{table*}[ht]
\centering
\caption{Performance on LongBench. KVSlimmer outperforms its baselines on most settings.}
\begin{tabular}{lccccccc}
\toprule
\textbf{} & {\textbf{Single-Doc}} & {\textbf{Multi-Doc}} & {\textbf{Sum}} & {\textbf{Few-shot} } & {\textbf{Synthetic}} & {\textbf{Code}} & \textbf{Avg.} \\



\midrule
\multicolumn{8}{c}{\textbf{Llama3.1-8B-Instruct}} \\
\midrule
Full Context & 43.73 & 44.49 & 29.12 & 69.36 & 53.56 & 50.95 & 48.07 \\
StreamingLLM & 28.15 & 27.19 & 25.15 & 63.17 & 16.33 & 52.15 & 35.50 \\
LongCache & 28.98 & 27.84 & 25.35 & 64.73 & 19.68 & 51.61 & 36.46 \\
H$_2$O & 33.30 & 34.43 & 26.60 & {\bf 66.23} & 14.75 & \textbf{52.65} & 38.53 \\
LLMLingua-2 & 32.02 & 32.24 & 24.99 & 27.87  & 17.67  & 50.07  & 30.43  \\
CaM & 32.14 & 32.63 & 24.91 &63.09 & 16.77 & 52.13 & 37.26 \\
{AsymKV} & {39.42} & {38.93} & {\textbf{27.30}} & {65.66} & {39.39} & {48.57} & {43.12} \\

{\textbf{KVSlimmer}} & {\textbf{40.24}} & {\textbf{39.61}} & {27.19} & {65.00} & {\textbf{44.52}} & {49.73} & {\textbf{44.04}} \\
\midrule
\multicolumn{8}{c}{\textbf{Mistral-7B-Instruct-v0.3}} \\
\midrule
Full Context & 38.74 & 38.29 & 29.04 & 70.70 & 51.00 & 53.05 & 46.15 \\
StreamingLLM & 24.80 & 22.14 & 25.18 & 66.49 & 15.14 & 52.10 & 34.39 \\
LongCache\textbf{ }& 26.05 & 22.31 & 25.44 & 66.21 & 14.93 & 51.03 & 34.50 \\
H$_2$O & 29.66 & 28.22 & 26.32 &  67.78& 14.83 & 51.63 & 36.80 \\
LLMLingua-2 & 28.12 & 28.62 & 25.75 & 45.85 & 16.00 & 46.50 & 31.88 \\
CaM\textbf{ }& 26.15 & 29.06 & 26.81 & 66.16 &20.96 & 51.40 & 36.83 \\
AsymKV & {\textbf{33.71}} & {\textbf{32.81}} & {\textbf{27.04}} & {67.21} & {34.56} & {51.33} & {40.88} \\
{\textbf{KVSlimmer}} & {33.42} & {32.62} & {26.83} & {\textbf{67.86}} & {\textbf{36.78}} & {\textbf{52.34}} & {\textbf{41.28}} \\
\midrule
\multicolumn{8}{c}{\textbf{Qwen2-1.5B-Instruct}} \\
\midrule
Full Context & 30.03 & 28.68 & 26.16 & 66.68 & 5.50 & 41.49 & 34.29 \\
StreamingLLM & 21.62 & 19.70 & 21.95 & 61.68 & 3.50 & 41.38 & 29.04 \\
LongCache & 22.22 & 27.31 & 16.28 & 56.65 & 5.25 & 41.21 & 28.76 \\
H$_2$O & 26.50 & 29.22 & 13.77 & 51.92 & 4.25 & 39.02 & 28.17 \\
LLMLingua-2 & 21.83 & 20.94 & 15.67 & 57.31 & 4.00 & 38.91 & 27.07 \\
CaM & 24.76 & 19.46 & 16.38 & 58.19 & 3.75 & 36.39 & 27.29 \\
{AsymKV} & 26.14 & 27.70  & 23.33  & {\textbf{62.99}} & 4.75  & 40.89  & 31.98 \\
{\textbf{KVSlimmer}} & {\textbf{26.54}} & {\textbf{29.54}} & {\textbf{23.90}} & {61.54} & {\textbf{5.50}} & {\textbf{41.84}} & {\textbf{32.45}} \\
%
\bottomrule
\end{tabular}
\label{tab:longbench}
\vskip -5pt
\end{table*}

\section{Experiments}

\subsection{Experimental Setup}
\textbf{Baselines.}
We compare KVSlimmer against several categories of approaches:
\emph{context segmentation}: StreamingLLM~\cite{attention_sink} and LongCache~\cite{liu2024farewell},
\emph{prompt compression}: LLMLingua-2.0~\cite{pan-etal-2024-llmlingua},
\emph{KV cache compression}: H$_2$O~\cite{h2o},
\emph{KV cache merge}: CaM~\cite{zhang2024cam} and AsymKV~\cite{asymkv}.

\textbf{Base Models.}
To demonstrate the generality of KVSlimmer, we conduct evaluations across a diverse set of model architectures, including 
Llama3.1-8B-Instruct~\cite{dubey2024llama}, Mistral-7B-Instruct-v0.3~\cite{jiang2023mistral7b} and Qwen2-1.5B-Instruct~\cite{yang2024qwen2technicalreport}.

\textbf{Implementation Details.}
Unless otherwise specified, we set the compression context budget to $2048$ tokens and the compression granularity chunk\_size to $512$. All baseline methods are evaluated under identical configurations to ensure fair comparison. For H$_2$O, we follow the original setup by allocating a recent-token budget of $2048$ and a heavy-token budget of $512$. Following Attention Sink~\cite{attention_sink}, the initial $32$ tokens are always preserved. All experiments are conducted on one NVIDIA A100 GPU with 80GB of memory.

\subsection{Long Context Performance Evaluation}
We evaluate KVSlimmer's effectiveness on LongBench~\cite{bai2024longbench}, a comprehensive long-context benchmark, containing 16 English tasks from a wide range of categories.

\textbf{Results.} As summarized in Table~\ref{tab:longbench}, KVSlimmer achieves the SOTA performance on LongBench, demonstrating consistent advantages across various model architectures and task categories.
%
%
Specifically, on Llama3.1-8B-Instruct, KVSlimmer reaches an average score of 44.04, surpassing the previous SOTA, AsymKV, by a margin of 0.92. Notably, the gains are particularly pronounced in long-context–sensitive tasks. For instance, it yields improvements of 0.82, 0.68, and 5.13 on Single-Doc, Multi-Doc, and Synthetic tasks, respectively.
%
%
Similarly, on Mistral-7B-Instruct-v0.3, KVSlimmer obtains the highest average score of 41.28, outperforming AsymKV by 0.40 and setting new benchmarks in Few-shot, Code, and Synthetic categories.
%
Even on the more compact Qwen2-1.5B-Instruct model, KVSlimmer maintains its lead with an average score of 32.45. This underscores its capability to effectively preserve critical information during KV merging, even when constrained by limited model capacity.

\begin{table*}[ht]
    \centering
    \caption{Performance on LongBenchV2.}
    \begin{tabular}{lcccccc}
        \hline
        \textbf{Model} & \textbf{Overall} & \textbf{Easy} & \textbf{Hard} & \textbf{Short} & \textbf{Medium} & \textbf{Long} \\
        \hline
        Full Context & 30.02 & 30.73 & 29.58 & 35.00 & 27.91 & 25.93 \\
        StreamingLLM & 27.04 & 27.60 & 26.69 & 32.78 & 23.26 & 25.00 \\
        LongCache & 28.43 & 28.13 & 28.62  & 32.78 & 25.58 & 26.85  \\
        H$_2$O   &  28.23  & 28.12 & 28.29 & 31.67 & 26.98 & 25.00\\
        CaM & 28.23 & 28.64 & 27.97 & 31.67 & 26.98 & 25.00\\
       AsymKV & 30.02 & 30.23 & \textbf{29.90} & 32.78& \textbf{27.44} & 28.85 \\
        \textbf{KVSlimmer} & \textbf{30.22} & \textbf{32.81} & 28.62 & \textbf{36.11} & 25.12 & \textbf{30.56} \\
        \hline
    \end{tabular}
    \label{tab:LongBenchV2}
\end{table*}

\textbf{Extreme Long-Context Compression.} 
We evaluate KVSlimmer on LongBenchV2~\cite{bai2024longbench2}, which comprises contexts ranging from 8K to 2M tokens across six task categories. The results, obtained using Llama3.1-8B-Instruct with a cache size of 8192, are reported in Table~\ref{tab:LongBenchV2}. Notably, KVSlimmer outperforms SOTA methods across Easy, Short, and Long categories, resulting in the best overall performance. These results demonstrate KVSlimmer's robustness and effectiveness in handling extremely long contexts under constrained cache budgets.

%


\begin{figure}[htbp]
  \centering
  \includegraphics[
    width=0.95\columnwidth]{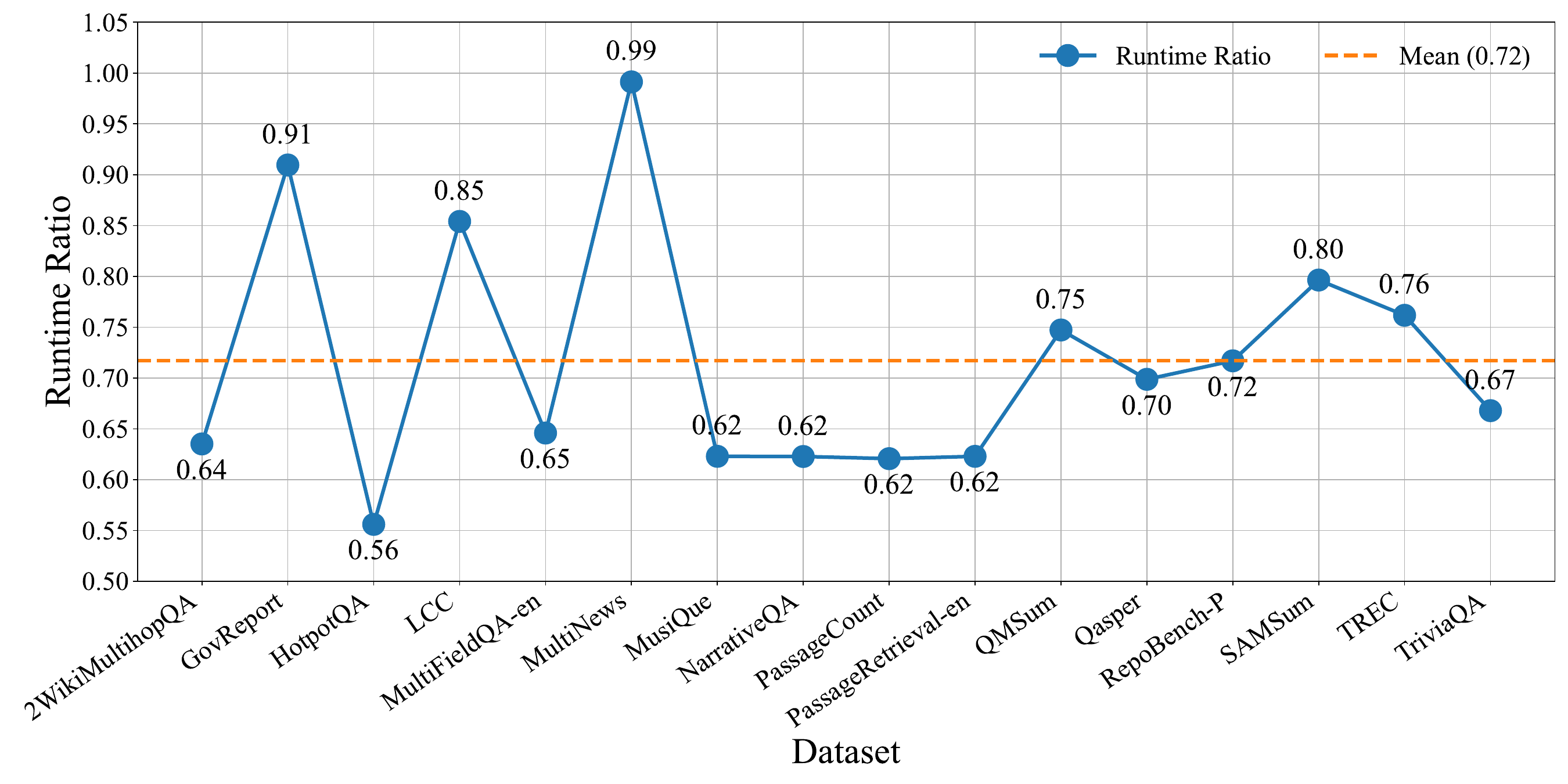}
  \caption{Relative runtime of KVSlimmer compared to AsymKV across LongBench datasets.}
  \label{fig:runtime_ratio}
\vskip -10pt
\end{figure}

\begin{figure}
    \centering  \includegraphics[width=0.55\linewidth]{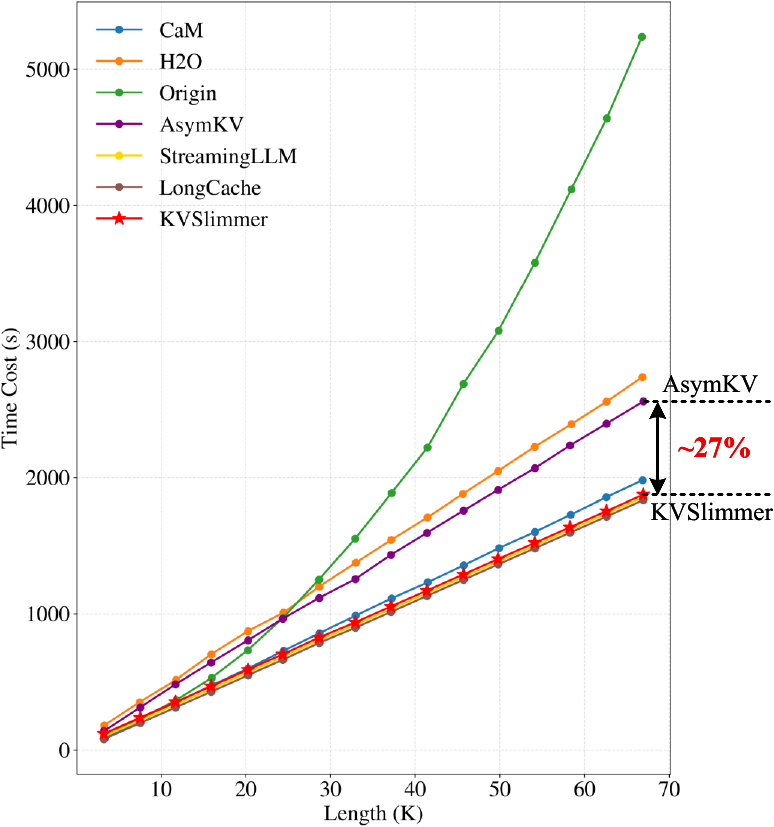}
    \caption{Inference efficiency of decoder stage.}
    \label{fig:inference efficiency}
    \vskip -5pt
\end{figure}

\begin{figure}[htbp]
  \centering
  \includegraphics[
    width=0.8\columnwidth,    
  ]{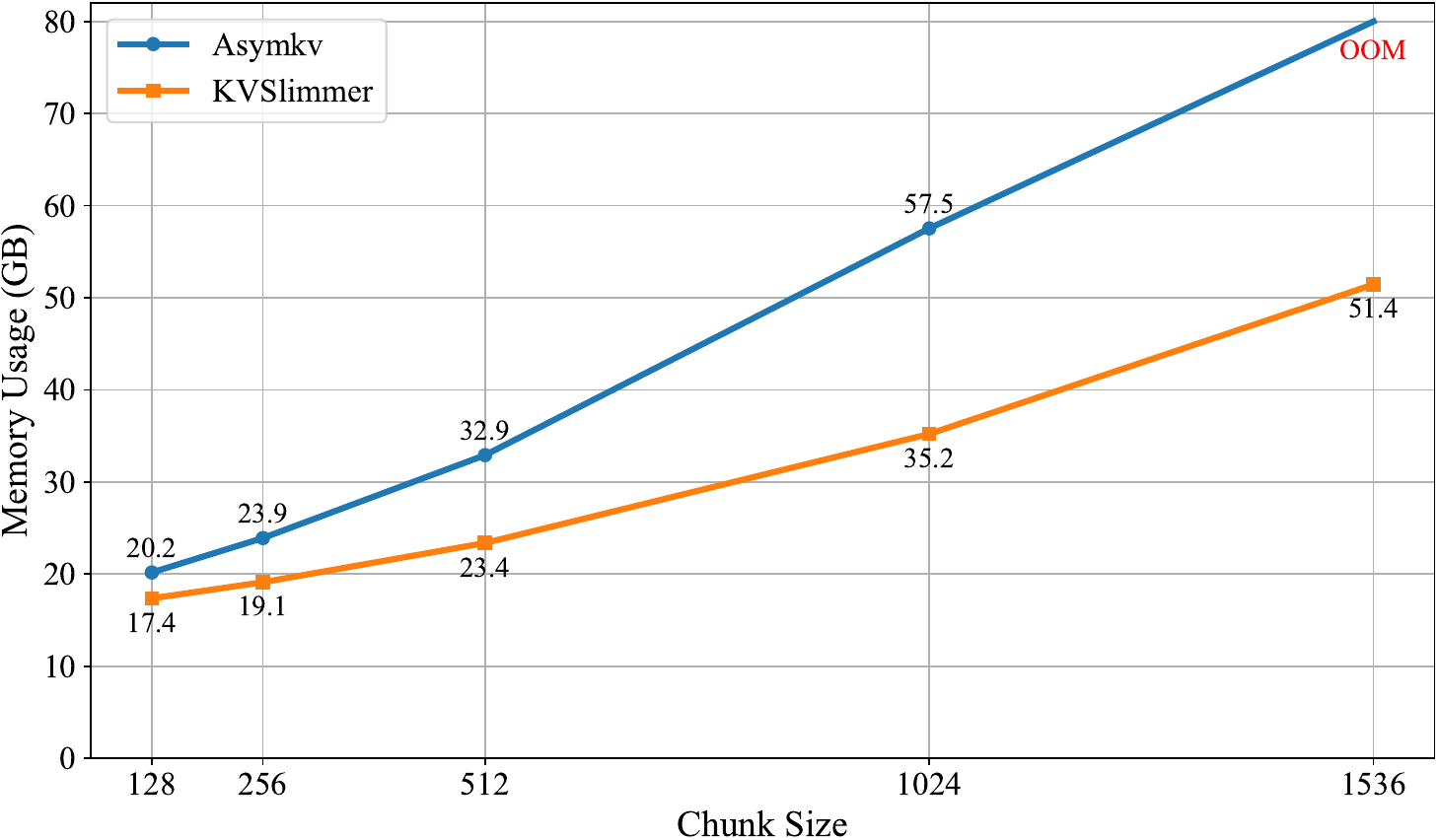}
  \caption{Peak GPU memory of KVSlimmer and AsymKV across different chunk sizes.}
  \label{fig:memory_usage_icml}
  \vskip -10pt
\end{figure}

\subsection{Runtime/Memory Efficiency}

Fig.~\ref{fig:runtime_ratio} illustrates the relative runtime of KVSlimmer compared with AsymKV across LongBench datasets. While KVSlimmer performs similarly to AsymKV on shorter tasks (\emph{e.g.}, GovReport and MultiNews), it demonstrates a substantial advantage in time efficiency as sequence length increases.
For instance, KVSlimmer achieves a 44\% reduction in runtime on long-context HotpotQA, and a 38\% reduction on MusiQue, NarrativeQA, PassageCount, and PassageRetrieval-en. Overall, KVSlimmer consistently reduces inference latency by an average of 28\%, highlighting its superior computational efficiency.
Fig.~\ref{fig:inference efficiency} compares the runtime of generating long-context sequences. As can be seen, thanks to the minimal KV cache operations, KVSlimmer reduces the time overhead much more significantly than AsymKV, and even yields comparable latency to context-segmentation approaches (\emph{e.g.}, StreamingLLM and LongCache). These results demonstrate that KVSlimmer strikes a better balance between efficiency and effectiveness.


%

Fig.~\ref{fig:memory_usage_icml} compares the peak GPU memory usage of KVSlimmer and AsymKV across various chunk sizes, averaged over all 16 LongBench tasks. While memory consumption for both methods scales with chunk size, KVSlimmer consistently maintains a significantly lower memory footprint. Notably, this efficiency gap widens as the chunk size grows. KVSlimmer achieves memory reductions of 29\% and 39\% at chunk sizes of 512 and 1024, respectively. These results demonstrate that KVSlimmer effectively mitigates memory pressure induced by large chunks, enabling more aggressive chunking strategies and more stable long-context inference within constrained GPU memory budgets.


\subsection{Compression Rate Analysis}

\begin{figure}[t]
  \centering
  \begin{subfigure}[t]{0.48\linewidth}
    \centering
    \includegraphics[width=\linewidth]{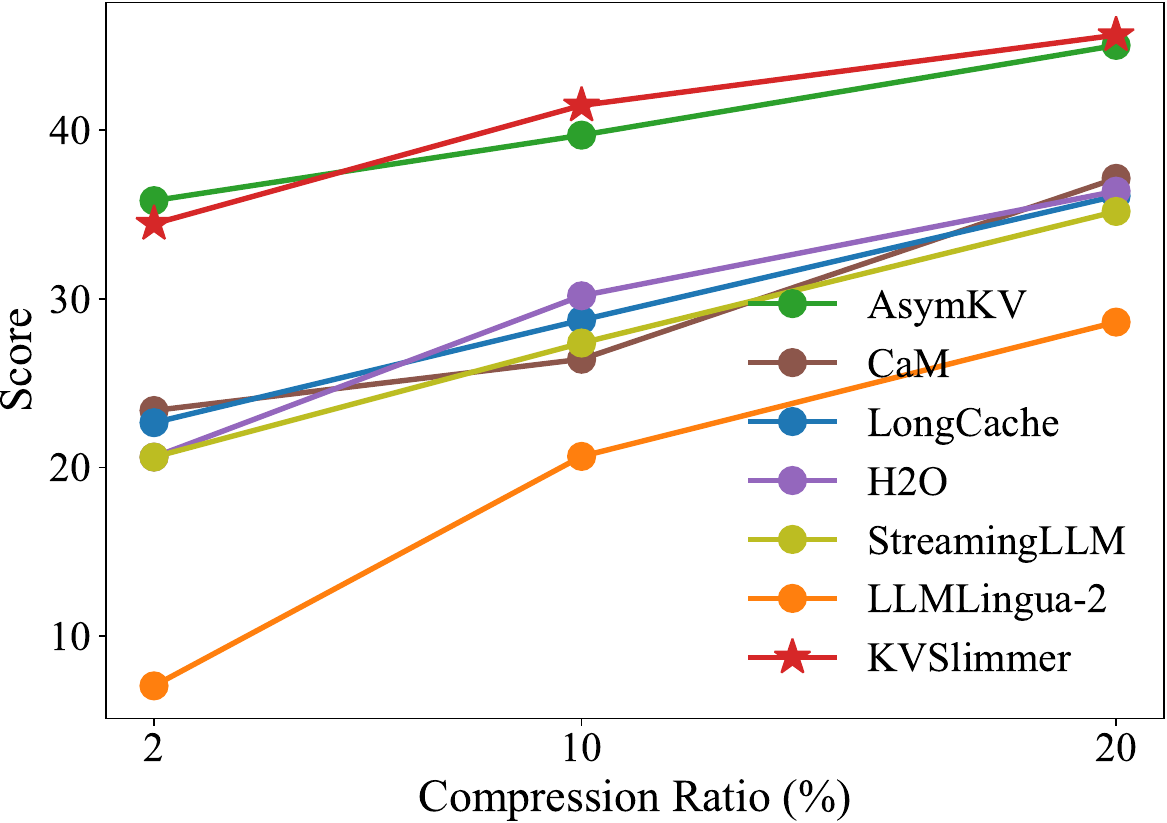}
    \caption{Mistral-7B.}
  \end{subfigure}
  \hfill
  \begin{subfigure}[t]{0.48\linewidth}
    \centering
    \includegraphics[width=\linewidth]{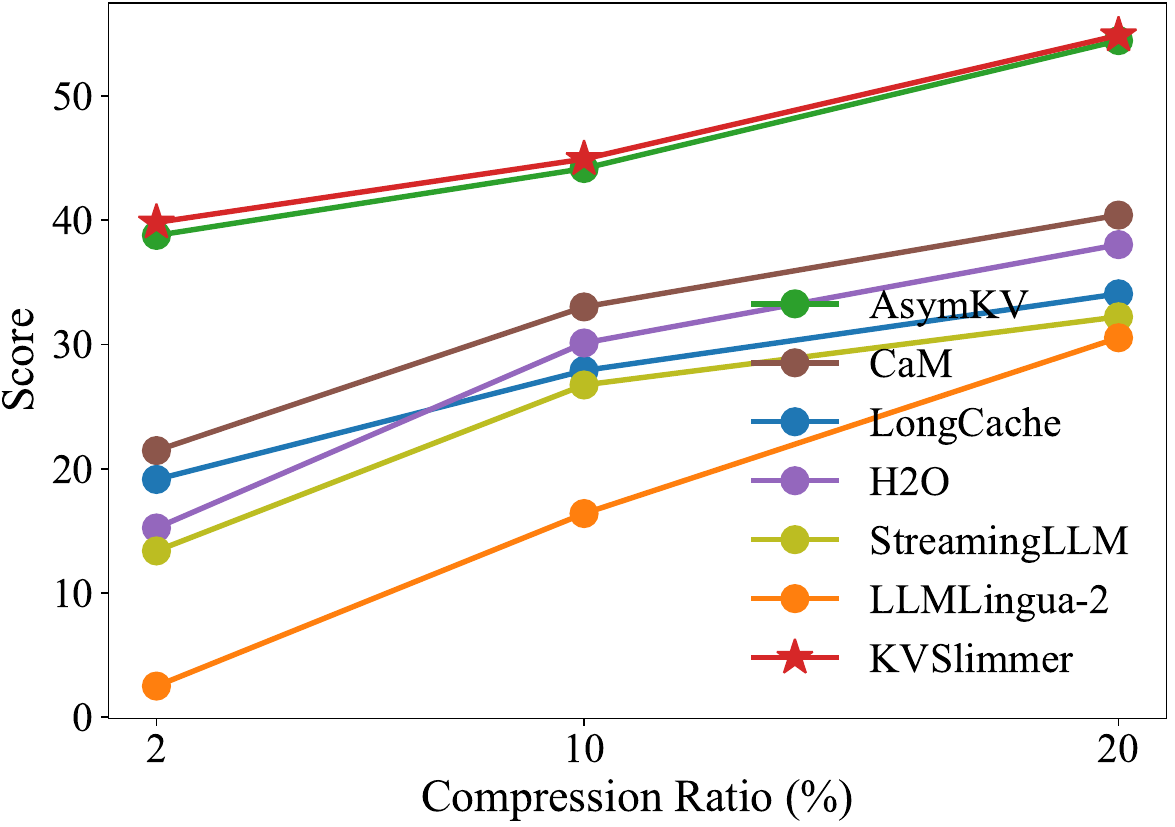}
    \caption{LLaMA-3.1-8B.}
  \end{subfigure}
  \caption{Results on different compression ratios.}
  \label{fig:longbench_mistral_llama}
\vskip -20pt
\end{figure}

To systematically compare the context compression capabilities of KVSlimmer and AsymKV, we analyze their performance across various compression ratios on the LongBench HotpotQA~\cite{yang2018hotpotqa} task. We define the compression ratio as the number of retained tokens relative to the original sequence length. As illustrated in Fig.~\ref{fig:longbench_mistral_llama}, KVSlimmer consistently achieves slightly superior performance compared to AsymKV on average. More importantly, KVSlimmer incurs significantly lower computational and memory overhead. For instance, at a 10\% compression ratio, it reduces latency and memory usage by $\sim$20\% and $\sim$27\%, respectively.


\section{Discussion}

Despite its theoretical and empirical advantages, KVSlimmer has several limitations that invite future exploration. 
%
%
First, our spectral analysis and merging strategy primarily focus on local token sequences. Exploring non-local or global merging could potentially yield even higher compression ratios by capturing long-range dependencies.
Second, although KVSlimmer is gradient-free and time-efficient, the current implementation employs a uniform compression ratio across all layers. Developing adaptive compression strategies that dynamically adjust the merging intensity based on the specific importance of each layer represents a promising direction for future research.

\section{Conclusion}

In this paper, we introduced KVSlimmer, a theoretically grounded and computationally efficient framework for asymmetric KV cache compression. By establishing a unified spectral analysis framework, we first unraveled the theoretical origins of QKV asymmetry in LLMs, demonstrating how the spectral energy distribution of projection weights dictates homogeneity and heterogeneity. 
Then, we introduce KVSlimmer, which derives a mathematically exact Hessian formulation that captures the off-diagonal coupling between adjacent Keys and achieves a gradient-free, closed-form solution that relies solely on forward-pass variables, reducing the memory and time overhead by a large margin. 
Extensive experiments on various LLMs and benchmarks validate that KVSlimmer significantly reduces memory and latency while maintaining, or even enhancing, model performance on long-context tasks. 

\section*{Impact Statement}

This paper presents work whose goal is to advance the field of Machine Learning. There are many potential societal consequences of our work, none of which we feel must be specifically highlighted here.

\bibliography{example_paper}

@inproceedings{NEURIPS2022_67d57c32,
 author = {Dao, Tri and Fu, Dan and Ermon, Stefano and Rudra, Atri and R\'{e}, Christopher},
 booktitle = {Advances in Neural Information Processing Systems},
 editor = {S. Koyejo and S. Mohamed and A. Agarwal and D. Belgrave and K. Cho and A. Oh},
 pages = {16344--16359},
 publisher = {Curran Associates, Inc.},
 title = {FlashAttention: Fast and Memory-Efficient Exact Attention with IO-Awareness},
 volume = {35},
 year = {2022}
}

@inproceedings{ijcai.2024/917,
author = {Wang, Xindi and Salmani, Mahsa and Omidi, Parsa and Ren, Xiangyu and Rezagholizadeh, Mehdi and Eshaghi, Armaghan},
title = {Beyond the limits: a survey of techniques to extend the context length in large language models},
year = {2024},
isbn = {978-1-956792-04-1},
doi = {10.24963/ijcai.2024/917},
booktitle = {Proceedings of the Thirty-Third International Joint Conference on Artificial Intelligence},
articleno = {917},
numpages = {9},
location = {Jeju, Korea},
series = {IJCAI '24}
}

@inproceedings{h2o,
 author = {Zhang, Zhenyu and Sheng, Ying and Zhou, Tianyi and Chen, Tianlong and Zheng, Lianmin and Cai, Ruisi and Song, Zhao and Tian, Yuandong and R\'{e}, Christopher and Barrett, Clark and Wang, Zhangyang "Atlas" and Chen, Beidi},
 booktitle = {Advances in Neural Information Processing Systems},
 editor = {A. Oh and T. Naumann and A. Globerson and K. Saenko and M. Hardt and S. Levine},
 pages = {34661--34710},
 publisher = {Curran Associates, Inc.},
 title = {H2O: Heavy-Hitter Oracle for Efficient Generative Inference of Large Language Models},

 volume = {36},
 year = {2023}
}

@inproceedings{Scissorhands,
 author = {Liu, Zichang and Desai, Aditya and Liao, Fangshuo and Wang, Weitao and Xie, Victor and Xu, Zhaozhuo and Kyrillidis, Anastasios and Shrivastava, Anshumali},
 booktitle = {Advances in Neural Information Processing Systems},
 editor = {A. Oh and T. Naumann and A. Globerson and K. Saenko and M. Hardt and S. Levine},
 pages = {52342--52364},
 publisher = {Curran Associates, Inc.},
 title = {Scissorhands: Exploiting the Persistence of Importance Hypothesis for LLM KV Cache Compression at Test Time},

 volume = {36},
 year = {2023}
}

@inproceedings{
ge2024model,
title={Model Tells You What to Discard: Adaptive {KV} Cache Compression for {LLM}s},
author={Suyu Ge and Yunan Zhang and Liyuan Liu and Minjia Zhang and Jiawei Han and Jianfeng Gao},
booktitle={The Twelfth International Conference on Learning Representations},
year={2024},

}

@inproceedings{
attention_sink,
title={Efficient Streaming Language Models with Attention Sinks},
author={Guangxuan Xiao and Yuandong Tian and Beidi Chen and Song Han and Mike Lewis},
booktitle={The Twelfth International Conference on Learning Representations},
year={2024},

}

@misc{
wang2025model,
title={Model Tells You Where to Merge: Adaptive {KV} Cache Merging for {LLM}s on Long-Context Tasks},
author={Zheng Wang and Boxiao Jin and Yuming Chang and Zhongzhi Yu and Minjia Zhang},
year={2025},

}

@inproceedings{
wan2025textdtexto,
title={$\mathrm{D}_2\mathrm{O}$
: Dynamic Discriminative Operations for Efficient Long-Context Inference of Large Language Models},
author={Zhongwei Wan and Xinjian Wu and Yu Zhang and Yi Xin and Chaofan Tao and Zhihong Zhu and Xin Wang and Siqi Luo and Jing Xiong and Longyue Wang and Mi Zhang},
booktitle={The Thirteenth International Conference on Learning Representations},
year={2025},

}

@inproceedings{
liu2024minicache,
title={MiniCache: {KV} Cache Compression in Depth Dimension for Large Language Models},
author={Akide Liu and Jing Liu and Zizheng Pan and Yefei He and Gholamreza Haffari and Bohan Zhuang},
booktitle={The Thirty-eighth Annual Conference on Neural Information Processing Systems},
year={2024},

}

@inproceedings{
asymkv,
title={Homogeneous Keys, Heterogeneous Values: Exploiting Local {KV} Cache Asymmetry for Long-Context {LLM}s},
author={Wanyun Cui and Mingwei Xu},
booktitle={The Thirty-ninth Annual Conference on Neural Information Processing Systems},
year={2025},

}

@inproceedings{
liu2023scissorhands,
title={Scissorhands: Exploiting the Persistence of Importance Hypothesis for {LLM} {KV} Cache Compression at Test Time},
author={Zichang Liu and Aditya Desai and Fangshuo Liao and Weitao Wang and Victor Xie and Zhaozhuo Xu and Anastasios Kyrillidis and Anshumali Shrivastava},
booktitle={Thirty-seventh Conference on Neural Information Processing Systems},
year={2023},
}

@inproceedings{10.5555/3737916.3738638,
author = {Li, Yuhong and Huang, Yingbing and Yang, Bowen and Venkitesh, Bharat and Locatelli, Acyr and Ye, Hanchen and Cai, Tianle and Lewis, Patrick and Chen, Deming},
title = {SnapKV: LLM knows what you are looking for before generation},
booktitle = {Advances in Neural Information Processing Systems 37},
year = {2024},
isbn = {9798331314385},
publisher = {Curran Associates Inc.},
address = {Red Hook, NY, USA},
articleno = {722},
numpages = {24},
location = {Vancouver, BC, Canada},
series = {NIPS '24}
}

@inproceedings{
ge2023model,
title={Model Tells You What to Discard: Adaptive {KV} Cache Compression for {LLM}s},
author={Suyu Ge and Yunan Zhang and Liyuan Liu and Minjia Zhang and Jiawei Han and Jianfeng Gao},
booktitle={Workshop on Advancing Neural Network Training: Computational Efficiency, Scalability, and Resource Optimization (WANT@NeurIPS 2023)},
year={2023},
}

@inproceedings{
cai2025pyramidkv,
title={Pyramid{KV}: Dynamic {KV} Cache Compression based on Pyramidal Information Funneling},
author={Zefan Cai and Yichi Zhang and Bofei Gao and Yuliang Liu and Yucheng Li and Tianyu Liu and Keming Lu and Wayne Xiong and Yue Dong and Junjie Hu and Wen Xiao},
booktitle={Second Conference on Language Modeling},
year={2025},

}

@inproceedings{
fu2025not,
title={Not All Heads Matter: A Head-Level {KV} Cache Compression Method with Integrated Retrieval and Reasoning},
author={Yu Fu and Zefan Cai and Abedelkadir Asi and Wayne Xiong and Yue Dong and Wen Xiao},
booktitle={The Thirteenth International Conference on Learning Representations},
year={2025},
}

@inproceedings{zhang2024cam,
title={CaM: Cache Merging for Memory-efficient {LLM}s Inference},
author={Yuxin Zhang and Yuxuan Du and Gen Luo and Yunshan Zhong and Zhenyu Zhang and Shiwei Liu and Rongrong Ji},
booktitle={Forty-first International Conference on Machine Learning},
year={2024},

}

@inproceedings{10.5555/3692070.3693589,
author = {Nawrot, Piotr and \L{}a\'{n}cucki, Adrian and Chochowski, Marcin and Tarjan, David and Ponti, Edoardo M.},
title = {Dynamic memory compression: retrofitting LLMs for accelerated inference},
year = {2024},
publisher = {JMLR.org},
booktitle = {Proceedings of the 41st International Conference on Machine Learning},
articleno = {1519},
numpages = {17},
location = {Vienna, Austria},
series = {ICML'24}
}

@inproceedings{dai-etal-2019-transformer,
    title = "Transformer-{XL}: Attentive Language Models beyond a Fixed-Length Context",
    author = "Dai, Zihang  and
      Yang, Zhilin  and
      Yang, Yiming  and
      Carbonell, Jaime  and
      Le, Quoc  and
      Salakhutdinov, Ruslan",
    editor = "Korhonen, Anna  and
      Traum, David  and
      M{\`a}rquez, Llu{\'i}s",
    booktitle = "Proceedings of the 57th Annual Meeting of the Association for Computational Linguistics",
    month = jul,
    year = "2019",
    address = "Florence, Italy",
    publisher = "Association for Computational Linguistics",
    
    doi = "10.18653/v1/P19-1285",
    pages = "2978--2988",
}

@inproceedings{
Rae2020Compressive,
title={Compressive Transformers for Long-Range Sequence Modelling},
author={Jack W. Rae and Anna Potapenko and Siddhant M. Jayakumar and Chloe Hillier and Timothy P. Lillicrap},
booktitle={International Conference on Learning Representations},
year={2020},

}

@inproceedings{10.5555/3600270.3601075,
author = {Bulatov, Aydar and Kuratov, Yuri and Burtsev, Mikhail S.},
title = {Recurrent memory transformer},
year = {2022},
isbn = {9781713871088},
publisher = {Curran Associates Inc.},
address = {Red Hook, NY, USA},
booktitle = {Proceedings of the 36th International Conference on Neural Information Processing Systems},
articleno = {805},
numpages = {13},
location = {New Orleans, LA, USA},
series = {NIPS '22}
}

@article{abs-2004-05150,
  publtype={informal},
  author={Iz Beltagy and Matthew E. Peters and Arman Cohan},
  title={Longformer: The Long-Document Transformer},
  year={2020},
  cdate={1577836800000},
  journal={CoRR},
  volume={abs/2004.05150},

}

@inproceedings{ZaheerGDAAOPRWY20,
  author={Manzil Zaheer and Guru Guruganesh and Kumar Avinava Dubey and Joshua Ainslie and Chris Alberti and Santiago Ontañón and Philip Pham and Anirudh Ravula and Qifan Wang and Li Yang and Amr Ahmed},
  title={Big Bird: Transformers for Longer Sequences},
  year={2020},
  cdate={1577836800000},
  
  booktitle={NeurIPS}
}

@article{abs-2402-17463,
  publtype={informal},
  author={Chenxin An and Fei Huang and Jun Zhang and Shansan Gong and Xipeng Qiu and Chang Zhou and Lingpeng Kong},
  title={Training-Free Long-Context Scaling of Large Language Models},
  year={2024},
  cdate={1704067200000},
  journal={CoRR},
  volume={abs/2402.17463},
 
}

@inproceedings{ainslie-etal-2020-etc,
    title = "{ETC}: Encoding Long and Structured Inputs in Transformers",
    author = "Ainslie, Joshua  and
      Onta{\~n}{\'o}n, Santiago  and
      Alberti, Chris  and
      Cvicek, Vaclav  and
      Fisher, Zachary  and
      Pham, Philip  and
      Ravula, Anirudh  and
      Sanghai, Sumit  and
      Wang, Qifan  and
      Yang, Li",
    editor = "Webber, Bonnie  and
      Cohn, Trevor  and
      He, Yulan  and
      Liu, Yang",
    booktitle = "Proceedings of the 2020 Conference on Empirical Methods in Natural Language Processing (EMNLP)",
    month = nov,
    year = "2020",
    address = "Online",
    publisher = "Association for Computational Linguistics",
   
    doi = "10.18653/v1/2020.emnlp-main.19",
    pages = "268--284"
}

@inproceedings{10.5555/3737916.3741717,
author = {Xiao, Chaojun and Zhang, Pengle and Han, Xu and Xiao, Guangxuan and Lin, Yankai and Zhang, Zhengyan and Liu, Zhiyuan and Sun, Maosong},
title = {InfLLM: training-free long-context extrapolation for LLMs with an efficient context memory},
booktitle = {Advances in Neural Information Processing Systems 37},
year = {2024},
isbn = {9798331314385},
publisher = {Curran Associates Inc.},
address = {Red Hook, NY, USA},
articleno = {3801},
numpages = {24},
location = {Vancouver, BC, Canada},
series = {NIPS '24}
}

@inproceedings{zhu-etal-2024-coca,
    title = "{C}o{CA}: Fusing Position Embedding with Collinear Constrained Attention in Transformers for Long Context Window Extending",
    author = "Zhu, Shiyi  and
      Ye, Jing  and
      Jiang, Wei  and
      Xue, Siqiao  and
      Zhang, Qi  and
      Wu, Yifan  and
      Li, Jianguo",
    editor = "Ku, Lun-Wei  and
      Martins, Andre  and
      Srikumar, Vivek",
    booktitle = "Proceedings of the 62nd Annual Meeting of the Association for Computational Linguistics (Volume 1: Long Papers)",
    month = aug,
    year = "2024",
    address = "Bangkok, Thailand",
    publisher = "Association for Computational Linguistics",
    
    doi = "10.18653/v1/2024.acl-long.233",
    pages = "4247--4262"
}

@inproceedings{wu-etal-2025-tokenselect,
    title = "{T}oken{S}elect: Efficient Long-Context Inference and Length Extrapolation for {LLM}s via Dynamic Token-Level {KV} Cache Selection",
    author = "Wu, Wei  and
      Pan, Zhuoshi  and
      Fu, Kun  and
      Wang, Chao  and
      Chen, Liyi  and
      Bai, Yunchu  and
      Wang, Tianfu  and
      Wang, Zheng  and
      Xiong, Hui",
    editor = "Christodoulopoulos, Christos  and
      Chakraborty, Tanmoy  and
      Rose, Carolyn  and
      Peng, Violet",
    booktitle = "Proceedings of the 2025 Conference on Empirical Methods in Natural Language Processing",
    month = nov,
    year = "2025",
    address = "Suzhou, China",
    publisher = "Association for Computational Linguistics",
   
    doi = "10.18653/v1/2025.emnlp-main.1079",
    pages = "21264--21281",
    ISBN = "979-8-89176-332-6"
}

@article{abs-2209-04881,
  publtype={informal},
  author={Feyza Duman Keles and Pruthuvi Mahesakya Wijewardena and Chinmay Hegde},
  title={On The Computational Complexity of Self-Attention},
  year={2022},
  cdate={1640995200000},
  journal={CoRR},
  volume={abs/2209.04881},

}

@article{liu2024farewell,
  title={Farewell to Length Extrapolation, a Training-Free Infinite Context with Finite Attention Scope},
  author={Liu, Xiaoran and Guo, Qipeng and Song, Yuerong and Liu, Zhigeng and Lv, Kai and Yan, Hang and Li, Linlin and Liu, Qun and Qiu, Xipeng},
  journal={arXiv preprint arXiv:2407.15176},
  year={2024}
}

@inproceedings{pan-etal-2024-llmlingua,
    title = "{LLML}ingua-2: Data Distillation for Efficient and Faithful Task-Agnostic Prompt Compression",
    author = "Zhuoshi Pan and Qianhui Wu and Huiqiang Jiang and Menglin Xia and Xufang Luo and Jue Zhang and Qingwei Lin and Victor Ruhle and Yuqing Yang and Chin-Yew Lin and H. Vicky Zhao and Lili Qiu and Dongmei Zhang",
    editor = "Ku, Lun-Wei  and
      Martins, Andre  and
      Srikumar, Vivek",
    booktitle = "Findings of the Association for Computational Linguistics ACL 2024",
    month = aug,
    year = "2024",
    address = "Bangkok, Thailand and virtual meeting",
    publisher = "Association for Computational Linguistics",
    
    pages = "963--981",
}

@article{dubey2024llama,
  title={The llama 3 herd of models},
  author={Dubey, Abhimanyu and Jauhri, Abhinav and Pandey, Abhinav and Kadian, Abhishek and Al-Dahle, Ahmad and Letman, Aiesha and Mathur, Akhil and Schelten, Alan and Yang, Amy and Fan, Angela and others},
  journal={arXiv preprint arXiv:2407.21783},
  year={2024}
}

@misc{jiang2023mistral7b,
      title={Mistral 7B}, 
      author={Albert Q. Jiang and Alexandre Sablayrolles and Arthur Mensch and Chris Bamford and Devendra Singh Chaplot and Diego de las Casas and Florian Bressand and Gianna Lengyel and Guillaume Lample and Lucile Saulnier and Lélio Renard Lavaud and Marie-Anne Lachaux and Pierre Stock and Teven Le Scao and Thibaut Lavril and Thomas Wang and Timothée Lacroix and William El Sayed},
      year={2023},
      eprint={2310.06825},
      archivePrefix={arXiv},
      primaryClass={cs.CL},
     
}

@inproceedings{bai2024longbench,
    title = "{L}ong{B}ench: A Bilingual, Multitask Benchmark for Long Context Understanding",
    author = "Bai, Yushi and Lv, Xin  and Zhang, Jiajie  and Lyu, Hongchang  and
      Tang, Jiankai  and Huang, Zhidian  and Du, Zhengxiao  and Liu, Xiao  and Zeng, Aohan  and Hou, Lei  and Dong, Yuxiao  and Tang, Jie  and Li, Juanzi",
    booktitle = "Proceedings of the 62nd Annual Meeting of the Association for Computational Linguistics (Volume 1: Long Papers)",
    month = aug,
    year = "2024",
    address = "Bangkok, Thailand",
    publisher = "Association for Computational Linguistics",
    
    doi = "10.18653/v1/2024.acl-long.172",
    pages = "3119--3137"
}

@inproceedings{bai2024longbench2,
  title={Longbench v2: Towards deeper understanding and reasoning on realistic long-context multitasks},
  author={Bai, Yushi and Tu, Shangqing and Zhang, Jiajie and Peng, Hao and Wang, Xiaozhi and Lv, Xin and Cao, Shulin and Xu, Jiazheng and Hou, Lei and Dong, Yuxiao and others},
  booktitle={Proceedings of the 63rd Annual Meeting of the Association for Computational Linguistics (Volume 1: Long Papers)},
  pages={3639--3664},
  year={2025}
}

@inproceedings{yang2018hotpotqa,
  title={HotpotQA: A Dataset for Diverse, Explainable Multi-hop Question Answering},
  author={Yang, Zhilin and Qi, Peng and Zhang, Saizheng and Bengio, Yoshua and Cohen, William and Salakhutdinov, Ruslan and Manning, Christopher D},
  booktitle={Proceedings of the 2018 Conference on Empirical Methods in Natural Language Processing},
  pages={2369--2380},
  year={2018}
}

@inproceedings{
ancucki2025inferencetime,
title={Inference-Time Hyper-Scaling with {KV} Cache Compression},
author={Adrian {\L}a{\'n}cucki and Konrad Staniszewski and Piotr Nawrot and Edoardo Ponti},
booktitle={The Thirty-ninth Annual Conference on Neural Information Processing Systems},
year={2025},

}

@inproceedings{
brandon2024reducing,
title={Reducing Transformer Key-Value Cache Size with Cross-Layer Attention},
author={William Brandon and Mayank Mishra and Aniruddha Nrusimha and Rameswar Panda and Jonathan Ragan-Kelley},
booktitle={The Thirty-eighth Annual Conference on Neural Information Processing Systems},
year={2024},

}

@misc{li2025flowmmcrossmodalinformationflow,
      title={FlowMM: Cross-Modal Information Flow Guided KV Cache Merging for Efficient Multimodal Context Inference}, 
      author={Kunxi Li and Yufan Xiong and Zhonghua Jiang and Yiyun Zhou and Zhaode Wang and Chengfei Lv and Shengyu Zhang},
      year={2025},
      eprint={2511.05534},
      archivePrefix={arXiv},
      primaryClass={cs.CL},
     
}

@misc{tian2025keepkvachievingperiodiclossless,
      title={KeepKV: Achieving Periodic Lossless KV Cache Compression for Efficient LLM Inference}, 
      author={Yuxuan Tian and Zihan Wang and Yebo Peng and Aomufei Yuan and Zhiming Wang and Bairen Yi and Xin Liu and Yong Cui and Tong Yang},
      year={2025},
      eprint={2504.09936},
      archivePrefix={arXiv},
      primaryClass={cs.LG},
    
}

@misc{chang2025xkvcrosslayersvdkvcache,
      title={xKV: Cross-Layer SVD for KV-Cache Compression}, 
      author={Chi-Chih Chang and Chien-Yu Lin and Yash Akhauri and Wei-Cheng Lin and Kai-Chiang Wu and Luis Ceze and Mohamed S. Abdelfattah},
      year={2025},
      eprint={2503.18893},
      archivePrefix={arXiv},
      primaryClass={cs.CL},
    
}

@misc{liu2025zsmergezeroshotkvcache,
      title={ZSMerge: Zero-Shot KV Cache Compression for Memory-Efficient Long-Context LLMs}, 
      author={Xin Liu and Xudong Wang and Pei Liu and Guoming Tang},
      year={2025},
      eprint={2503.10714},
      archivePrefix={arXiv},
      primaryClass={cs.CL},
    
}

@misc{yuan2025weightedkvattentionscoresweighted,
      title={WeightedKV: Attention Scores Weighted Key-Value Cache Merging for Large Language Models}, 
      author={Jian Yuan and Ziwei He and Haoli Bai and Jingwen Leng and Bo Jiang},
      year={2025},
      eprint={2503.01330},
      archivePrefix={arXiv},
      primaryClass={cs.CL},
     
}

@misc{gu2025obcacheoptimalbrainkv,
      title={OBCache: Optimal Brain KV Cache Pruning for Efficient Long-Context LLM Inference}, 
      author={Yuzhe Gu and Xiyu Liang and Jiaojiao Zhao and Enmao Diao},
      year={2025},
      eprint={2510.07651},
      archivePrefix={arXiv},
      primaryClass={cs.CL},
    
}

@misc{kim2025epicacheepisodickvcache,
      title={EpiCache: Episodic KV Cache Management for Long Conversational Question Answering}, 
      author={Minsoo Kim and Arnav Kundu and Han-Byul Kim and Richa Dixit and Minsik Cho},
      year={2025},
      eprint={2509.17396},
      archivePrefix={arXiv},
      primaryClass={cs.CL},
     
}

@misc{chittyvenkata2025pagedevictionstructuredblockwisekv,
      title={PagedEviction: Structured Block-wise KV Cache Pruning for Efficient Large Language Model Inference}, 
      author={Krishna Teja Chitty-Venkata and Jie Ye and Xian-He Sun and Anthony Kougkas and Murali Emani and Venkatram Vishwanath and Bogdan Nicolae},
      year={2025},
      eprint={2509.04377},
      archivePrefix={arXiv},
      primaryClass={cs.LG},
     
}

@misc{wang2025lookaheadqcacheachievingconsistent,
      title={Lookahead Q-Cache: Achieving More Consistent KV Cache Eviction via Pseudo Query}, 
      author={Yixuan Wang and Shiyu Ji and Yijun Liu and Yuzhuang Xu and Yang Xu and Qingfu Zhu and Wanxiang Che},
      year={2025},
      eprint={2505.20334},
      archivePrefix={arXiv},
      primaryClass={cs.CL},
     
}

@misc{gu2025ahakvadaptiveholisticattentiondriven,
      title={AhaKV: Adaptive Holistic Attention-Driven KV Cache Eviction for Efficient Inference of Large Language Models}, 
      author={Yifeng Gu and Zicong Jiang and Jianxiu Jin and Kailing Guo and Ziyang Zhang and Xiangmin Xu},
      year={2025},
      eprint={2506.03762},
      archivePrefix={arXiv},
      primaryClass={cs.CL},
      
}

@misc{feng2025evicpressjointkvcachecompression,
      title={EVICPRESS: Joint KV-Cache Compression and Eviction for Efficient LLM Serving}, 
      author={Shaoting Feng and Yuhan Liu and Hanchen Li and Xiaokun Chen and Samuel Shen and Kuntai Du and Zhuohan Gu and Rui Zhang and Yuyang Huang and Yihua Cheng and Jiayi Yao and Qizheng Zhang and Ganesh Ananthanarayanan and Junchen Jiang},
      year={2025},
      eprint={2512.14946},
      archivePrefix={arXiv},
      primaryClass={cs.OS},
      
}

@misc{feng2025tamingfragilitykvcache,
      title={Taming the Fragility of KV Cache Eviction in LLM Inference}, 
      author={Yuan Feng and Haoyu Guo and JunLin Lv and S. Kevin Zhou and Xike Xie},
      year={2025},
      eprint={2510.13334},
      archivePrefix={arXiv},
      primaryClass={cs.CL},
     
}

@misc{li2025emsadaptiveevictthenmergestrategy,
      title={EMS: Adaptive Evict-then-Merge Strategy for Head-wise KV Cache Compression Based on Global-Local Importance}, 
      author={Yingxin Li and Ye Li and Yuan Meng and Xinzhu Ma and Zihan Geng and Shutao Xia and Zhi Wang},
      year={2025},
      eprint={2412.08521},
      archivePrefix={arXiv},
      primaryClass={cs.CL},
     
}

@misc{wang2024modeltellsmergeadaptive,
      title={Model Tells You Where to Merge: Adaptive KV Cache Merging for LLMs on Long-Context Tasks}, 
      author={Zheng Wang and Boxiao Jin and Zhongzhi Yu and Minjia Zhang},
      year={2024},
      eprint={2407.08454},
      archivePrefix={arXiv},
      primaryClass={cs.CL},
   
}

@misc{liu2025surveytransformercontextextension,
      title={A Survey on Transformer Context Extension: Approaches and Evaluation}, 
      author={Yijun Liu and Jinzheng Yu and Yang Xu and Zhongyang Li and Qingfu Zhu},
      year={2025},
      eprint={2503.13299},
      archivePrefix={arXiv},
      primaryClass={cs.CL},
      url={https://arxiv.org/abs/2503.13299}, 
}

@misc{huang2024advancingtransformerarchitecturelongcontext,
      title={Advancing Transformer Architecture in Long-Context Large Language Models: A Comprehensive Survey}, 
      author={Yunpeng Huang and Jingwei Xu and Junyu Lai and Zixu Jiang and Taolue Chen and Zenan Li and Yuan Yao and Xiaoxing Ma and Lijuan Yang and Hao Chen and Shupeng Li and Penghao Zhao},
      year={2024},
      eprint={2311.12351},
      archivePrefix={arXiv},
      primaryClass={cs.CL},
      url={https://arxiv.org/abs/2311.12351}, 
}

@misc{liu2025spakelongcontextlargelanguage,
      title={Thus Spake Long-Context Large Language Model}, 
      author={Xiaoran Liu and Ruixiao Li and Mianqiu Huang and Zhigeng Liu and Yuerong Song and Qipeng Guo and Siyang He and Qiqi Wang and Linlin Li and Qun Liu and Ziwei He and Yaqian Zhou and Xuanjing Huang and Xipeng Qiu},
      year={2025},
      eprint={2502.17129},
      archivePrefix={arXiv},
      primaryClass={cs.CL},
      url={https://arxiv.org/abs/2502.17129}, 
}

@misc{yang2024qwen2technicalreport,
      title={Qwen2 Technical Report}, 
      author={An Yang and Baosong Yang and Binyuan Hui and Bo Zheng and Bowen Yu and Chang Zhou and Chengpeng Li and Chengyuan Li and Dayiheng Liu and Fei Huang and Guanting Dong and Haoran Wei and Huan Lin and Jialong Tang and Jialin Wang and Jian Yang and Jianhong Tu and Jianwei Zhang and Jianxin Ma and Jianxin Yang and Jin Xu and Jingren Zhou and Jinze Bai and Jinzheng He and Junyang Lin and Kai Dang and Keming Lu and Keqin Chen and Kexin Yang and Mei Li and Mingfeng Xue and Na Ni and Pei Zhang and Peng Wang and Ru Peng and Rui Men and Ruize Gao and Runji Lin and Shijie Wang and Shuai Bai and Sinan Tan and Tianhang Zhu and Tianhao Li and Tianyu Liu and Wenbin Ge and Xiaodong Deng and Xiaohuan Zhou and Xingzhang Ren and Xinyu Zhang and Xipin Wei and Xuancheng Ren and Xuejing Liu and Yang Fan and Yang Yao and Yichang Zhang and Yu Wan and Yunfei Chu and Yuqiong Liu and Zeyu Cui and Zhenru Zhang and Zhifang Guo and Zhihao Fan},
      year={2024},
      eprint={2407.10671},
      archivePrefix={arXiv},
      primaryClass={cs.CL},
      
}
\bibliographystyle{icml2026}

\newpage
\appendix
\onecolumn
 
\section{More Illustrations of Layer-wise QKV Similarity and Spectral Analysis}
\label{A}

\begin{figure*}
  \centering
  \includegraphics[
    width=\textwidth]{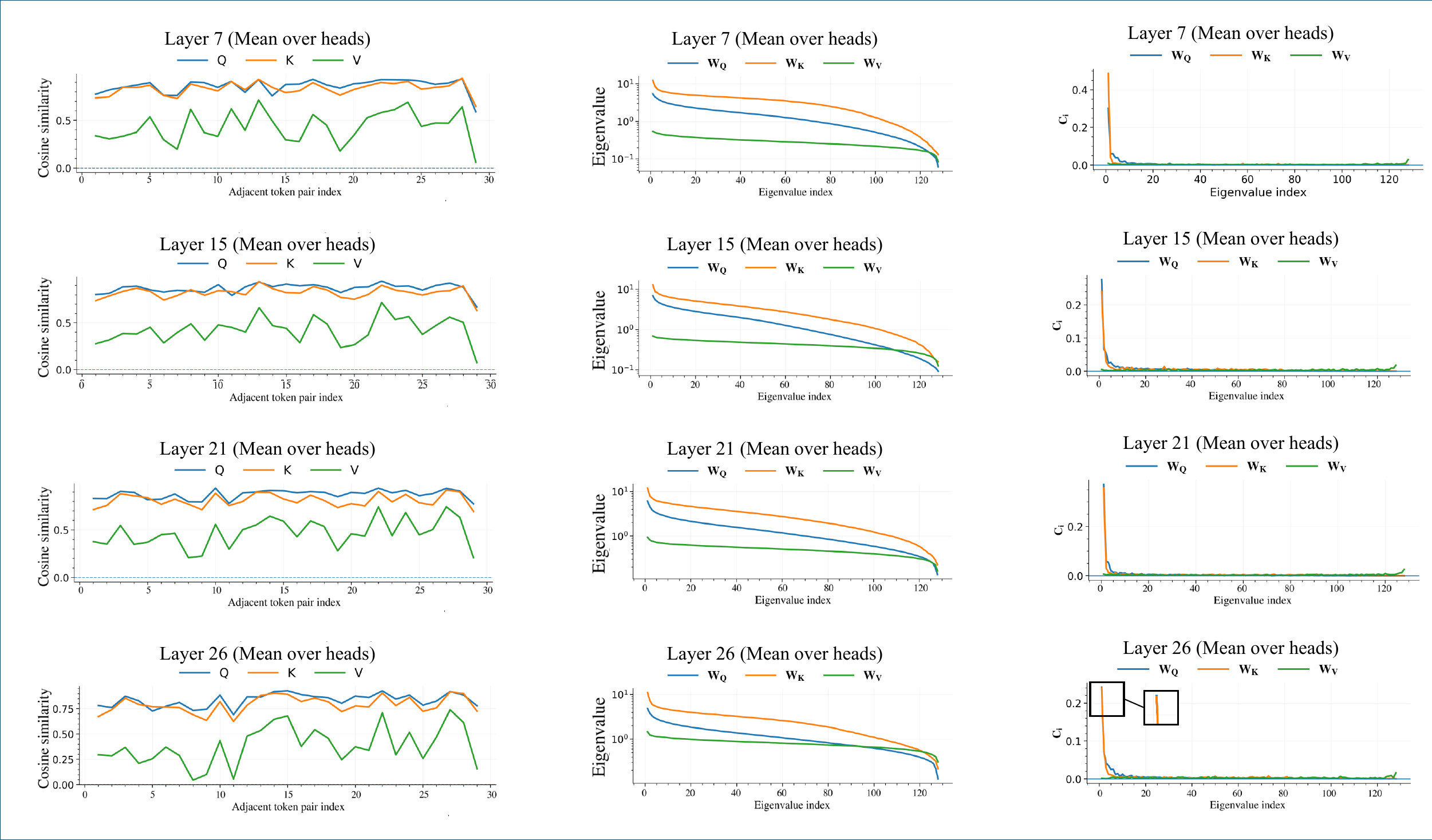}
  \caption{Layer-wise QKV similarity and spectral analysis for \textit{Llama-3.1-8B-Instruct}.
  Left column: Mean adjacent-token cosine similarity for Query (Q), Key (K), and Value (V), averaged over attention heads.
  Middle column: Eigenvalue distributions of the projection matrices $\mathbf{W}_Q$, $\mathbf{W}_K$, and $\mathbf{W}_V$, sorted in descending order.
  Right column: Mode-wise contribution coefficients $c_i$ (Eq.~\ref{eq:mode_contribution}), plotted according to the eigenvalue index.}
  \label{fig:Llama-3.1-8B-Instruct}
\end{figure*}

\begin{figure*}
  \centering
  \includegraphics[
    width=\textwidth]{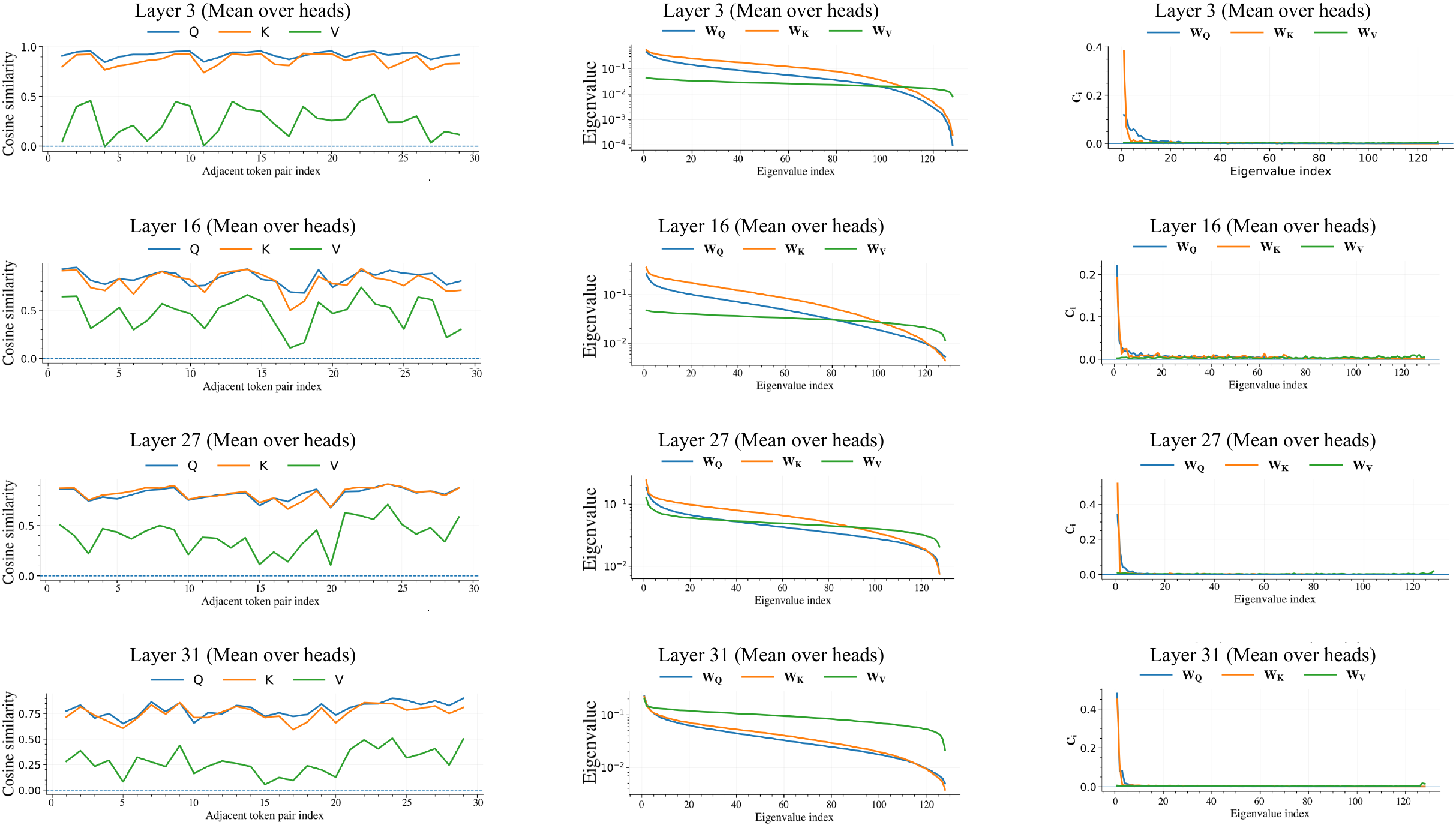}
  \caption{Layer-wise QKV similarity and spectral analysis for \textit{Mistral-7B-Instruct-v0.3}.
  Left column: Mean adjacent-token cosine similarity for Query (Q), Key (K), and Value (V), averaged over attention heads.
  Middle column: Eigenvalue distributions of the projection matrices $\mathbf{W}_Q$, $\mathbf{W}_K$, and $\mathbf{W}_V$, sorted in descending order.
  Right column: Mode-wise contribution coefficients $c_i$ (Eq.~\ref{eq:mode_contribution}), plotted according to the eigenvalue index.}
  \label{fig:Mistral-7B-Instruct}
\end{figure*}

In Fig.~\ref{fig:Llama-3.1-8B-Instruct} and Fig.~\ref{fig:Mistral-7B-Instruct}, we provide more layer-wise QKV similarity and spectral analysis across different models.





\section{Head-Level Alignment Illustrations}
\label{B}

\begin{figure} [htbp]
    \centering
    \begin{subfigure}{0.32\linewidth}
        \centering
        \includegraphics[width=\linewidth]{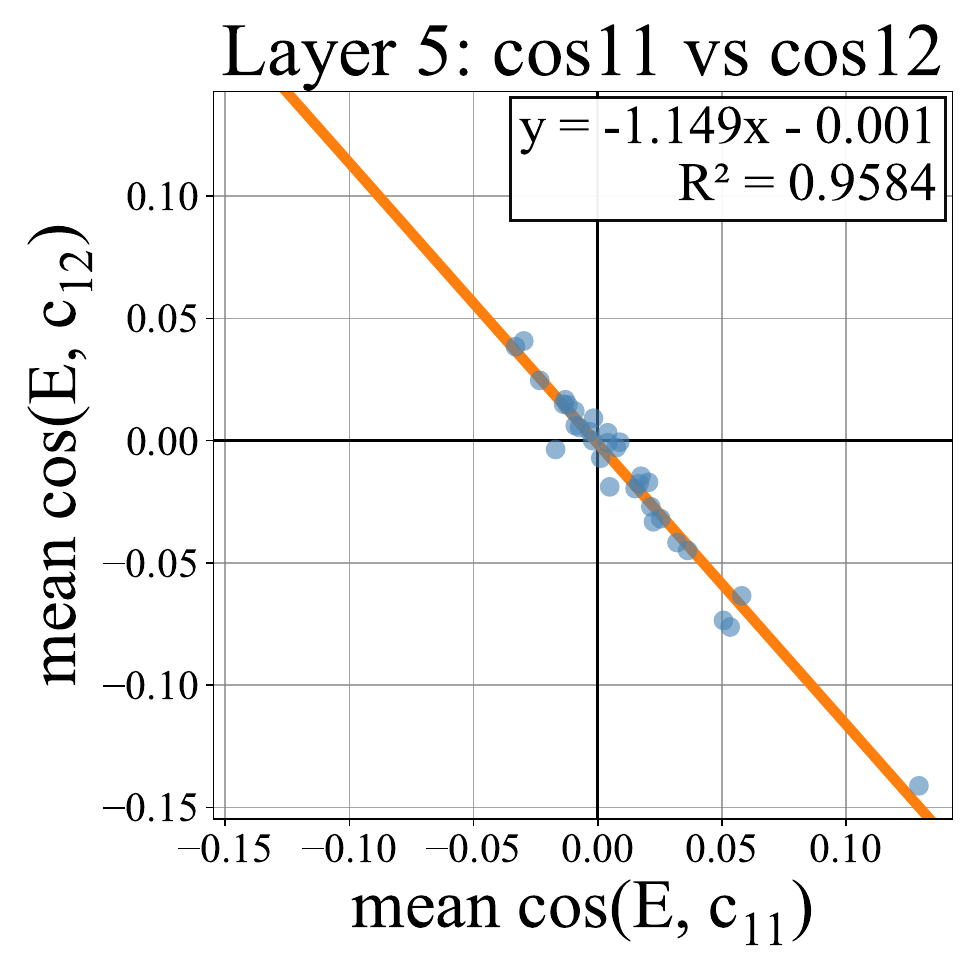}
        \caption{$\cos(\mathbf{E},\mathbf{c}_{11})$ vs. $\cos(\mathbf{E},\mathbf{c}_{12})$}
    \end{subfigure}\hfill
    \begin{subfigure}{0.32\linewidth}
        \centering
        \includegraphics[width=\linewidth]{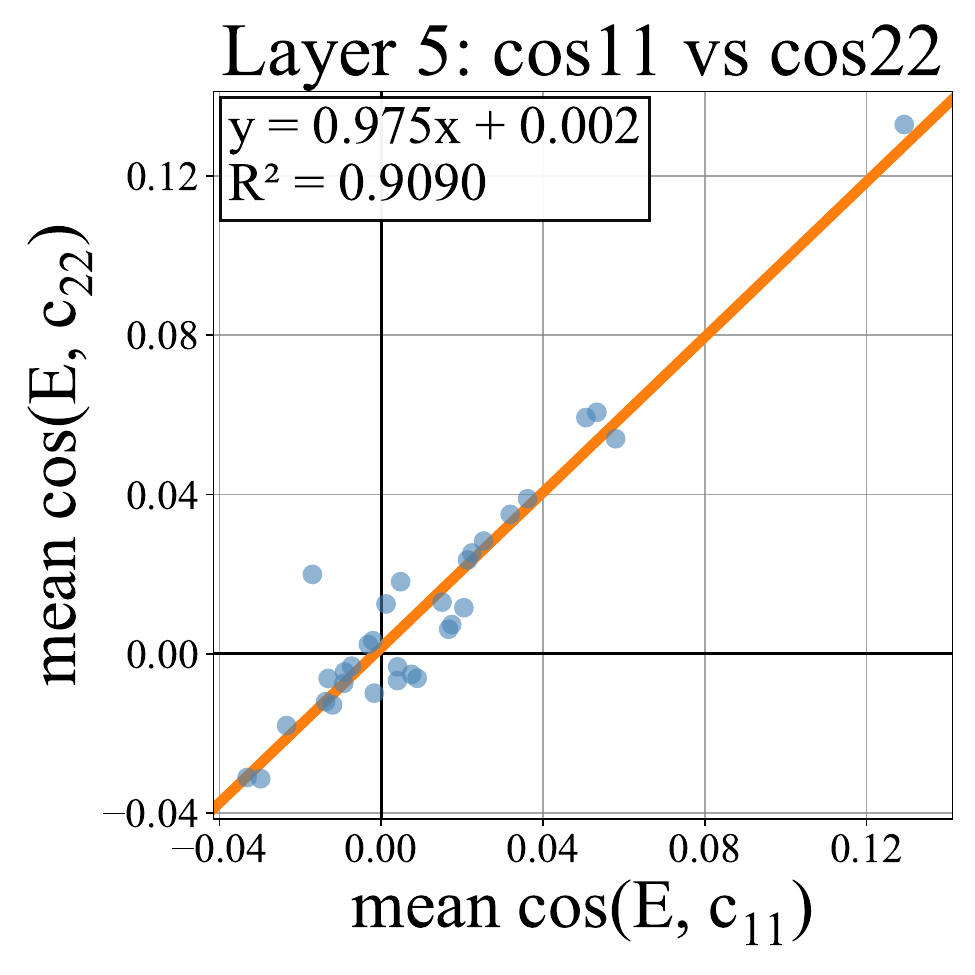}
        \caption{$\cos(\mathbf{E},\mathbf{c}_{11})$ vs. $\cos(\mathbf{E},\mathbf{c}_{22})$}
    \end{subfigure}\hfill
    \begin{subfigure}{0.32\linewidth}
        \centering
        \includegraphics[width=\linewidth]{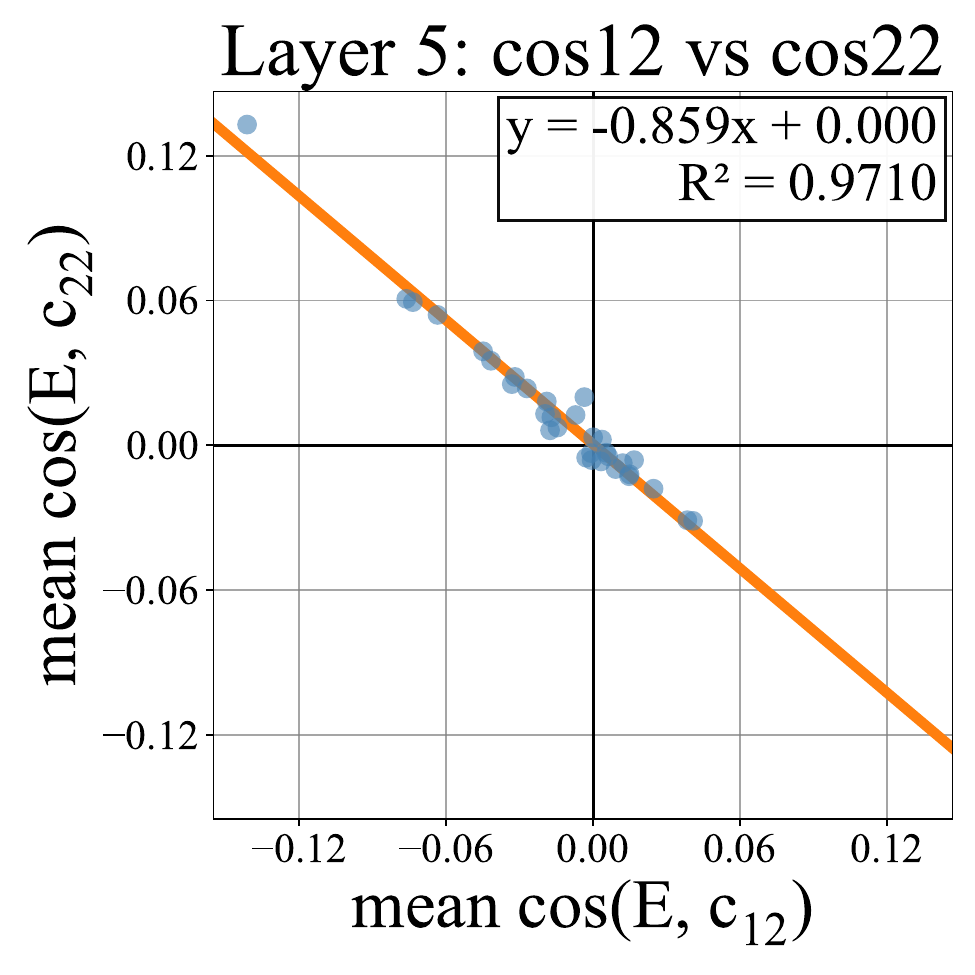}
        \caption{$\cos(\mathbf{E},\mathbf{c}_{12})$ vs. $\cos(\mathbf{E},\mathbf{c}_{22})$}
    \end{subfigure}

    \caption{Head-level mean alignment relationships at Layer~5 of \textit{Llama-3.1-8B-Instruct} on 2WikiMQA.
Each point corresponds to one attention head, positioned by its global mean cosine alignment.}
    \label{llama/Layer5}
\end{figure}

\begin{figure}
    \centering
    \begin{subfigure}{0.32\linewidth}
        \centering
        \includegraphics[width=\linewidth]{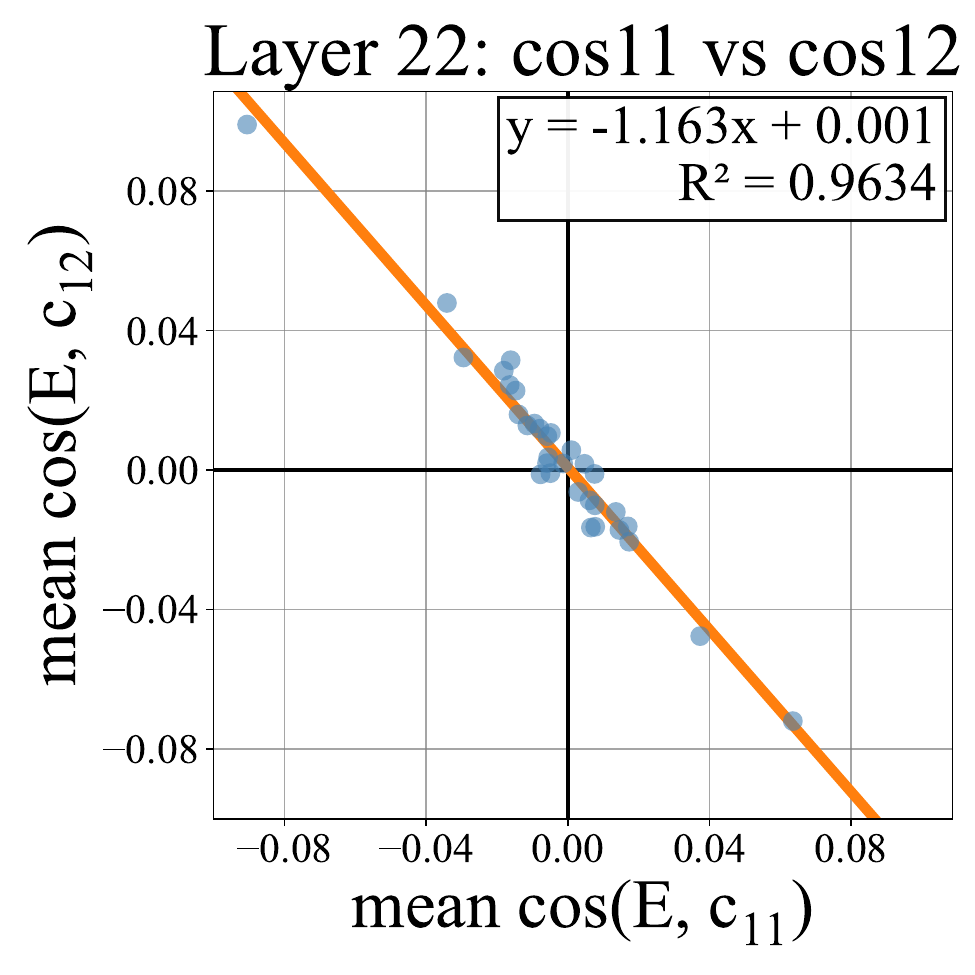}
        \caption{$\cos(\mathbf{E},\mathbf{c}_{11})$ vs. $\cos(\mathbf{E},\mathbf{c}_{12})$}
    \end{subfigure}\hfill
    \begin{subfigure}{0.32\linewidth}
        \centering
        \includegraphics[width=\linewidth]{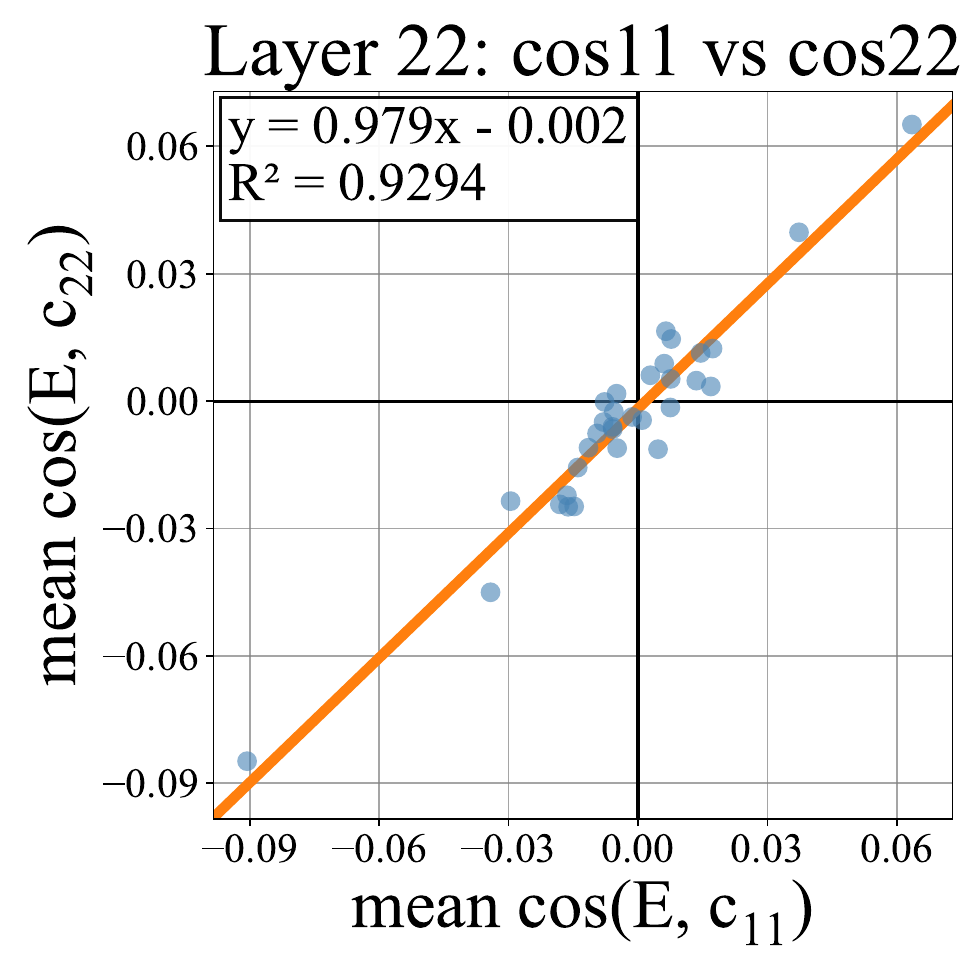}
        \caption{$\cos(\mathbf{E},\mathbf{c}_{11})$ vs. $\cos(\mathbf{E},\mathbf{c}_{22})$}
    \end{subfigure}\hfill
    \begin{subfigure}{0.32\linewidth}
        \centering
        \includegraphics[width=\linewidth]{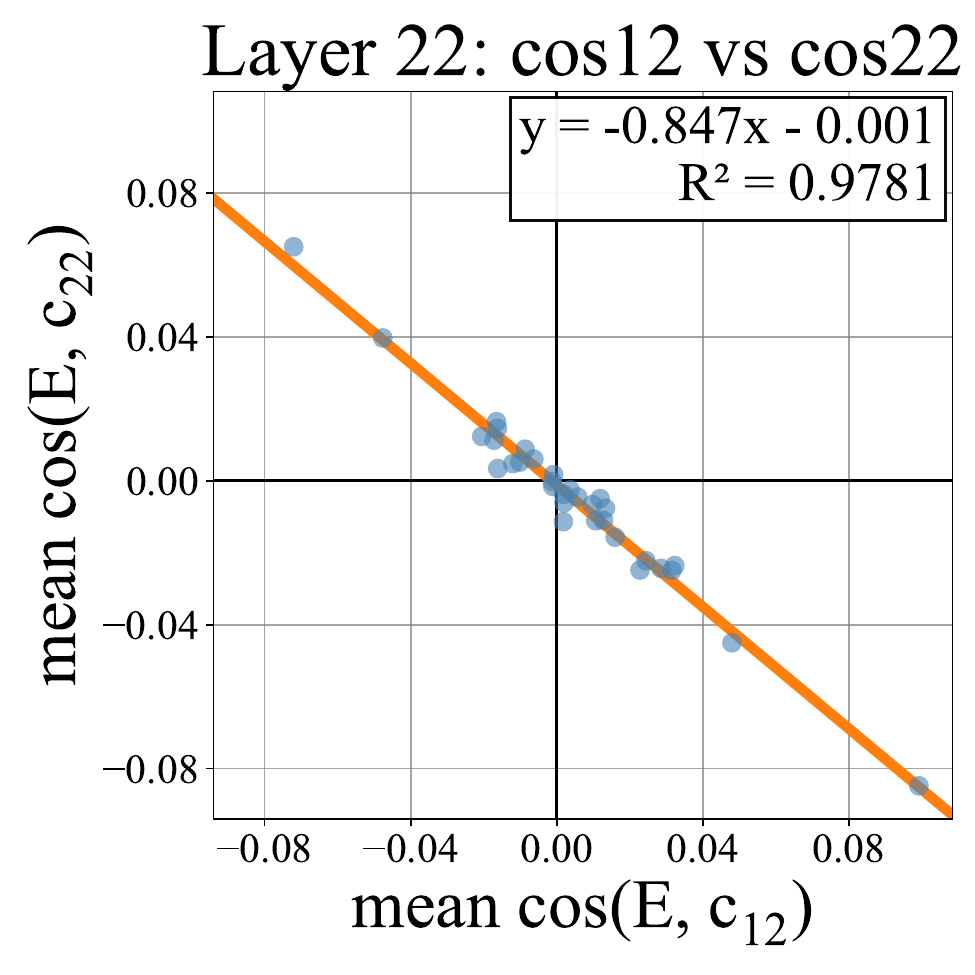}
        \caption{$\cos(\mathbf{E},\mathbf{c}_{12})$ vs. $\cos(\mathbf{E},\mathbf{c}_{22})$}
    \end{subfigure}

    \caption{Head-level mean alignment relationships at Layer~22 of \textit{Llama-3.1-8B-Instruct} on 2WikiMQA.
Each point corresponds to one attention head, positioned by its global mean cosine alignment.}
    \label{llama/Layer22}
\end{figure}

\begin{figure}
    \centering
    \begin{subfigure}{0.32\linewidth}
        \centering
        \includegraphics[width=\linewidth]{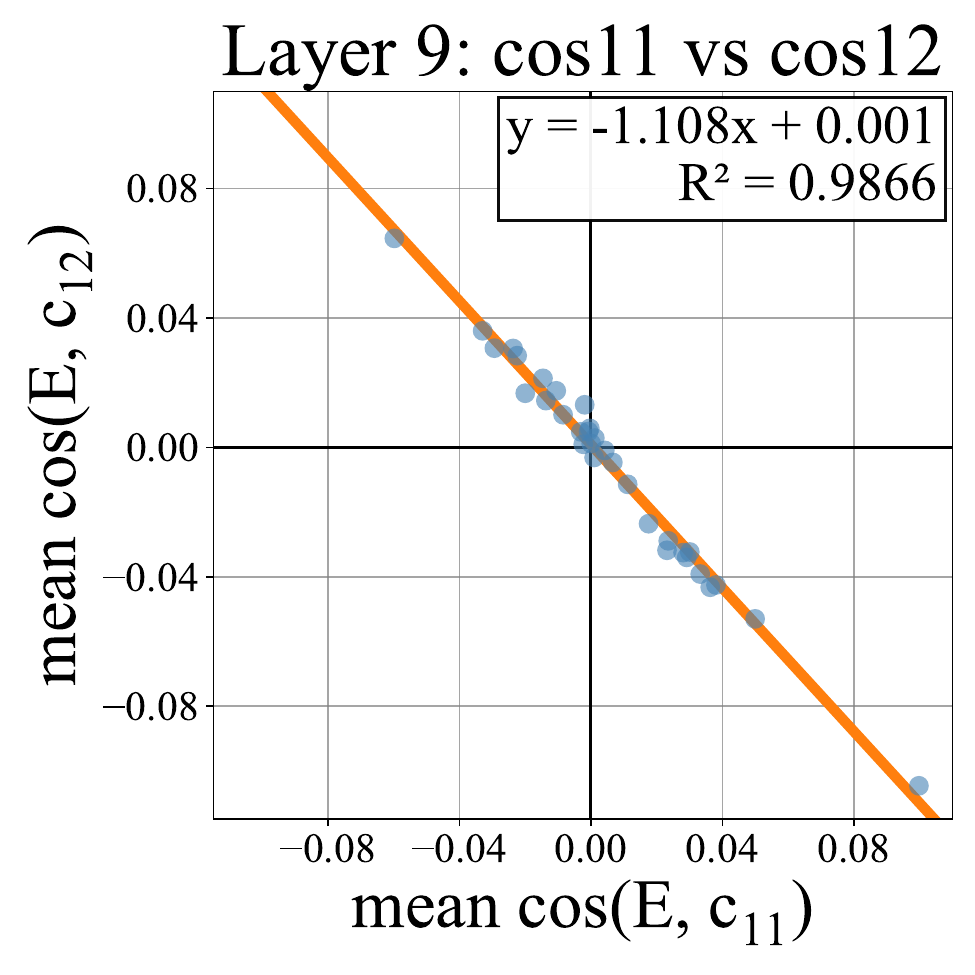}
        \caption{$\cos(\mathbf{E},\mathbf{c}_{11})$ vs. $\cos(\mathbf{E},\mathbf{c}_{12})$}
    \end{subfigure}\hfill
    \begin{subfigure}{0.32\linewidth}
        \centering
        \includegraphics[width=\linewidth]{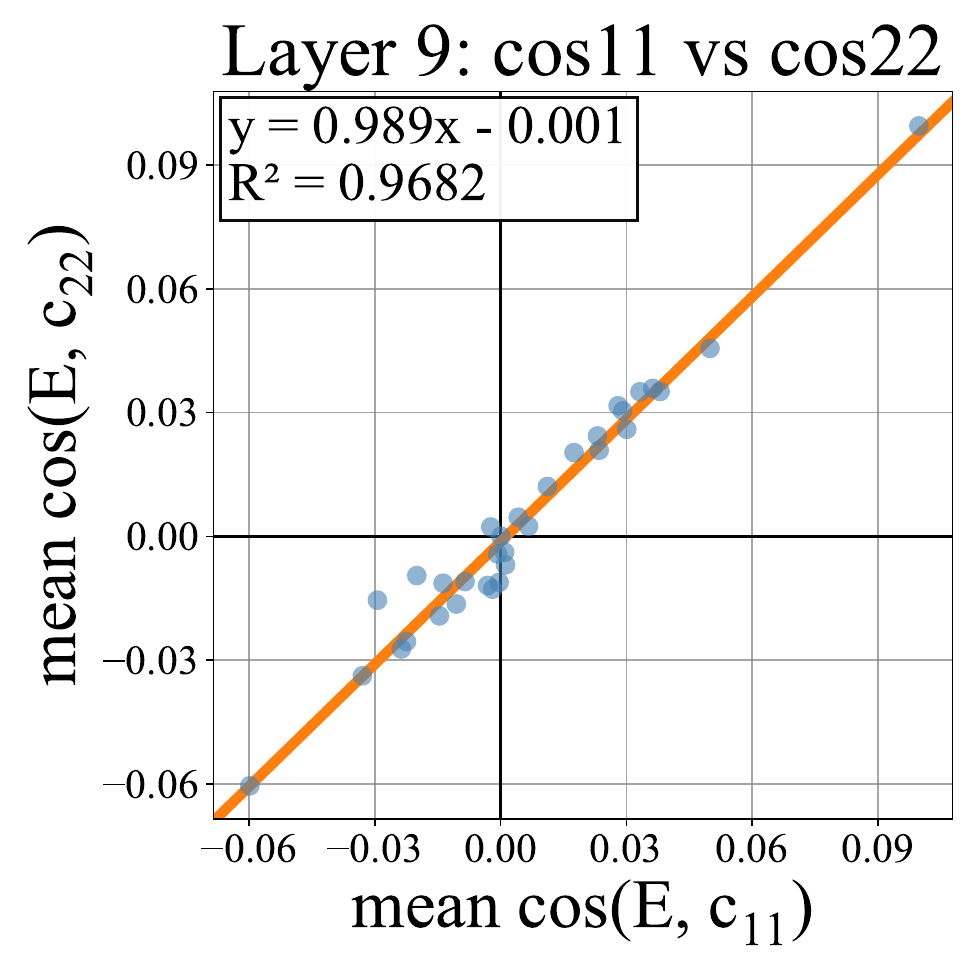}
        \caption{$\cos(\mathbf{E},\mathbf{c}_{11})$ vs. $\cos(\mathbf{E},\mathbf{c}_{22})$}
    \end{subfigure}\hfill
    \begin{subfigure}{0.32\linewidth}
        \centering
        \includegraphics[width=\linewidth]{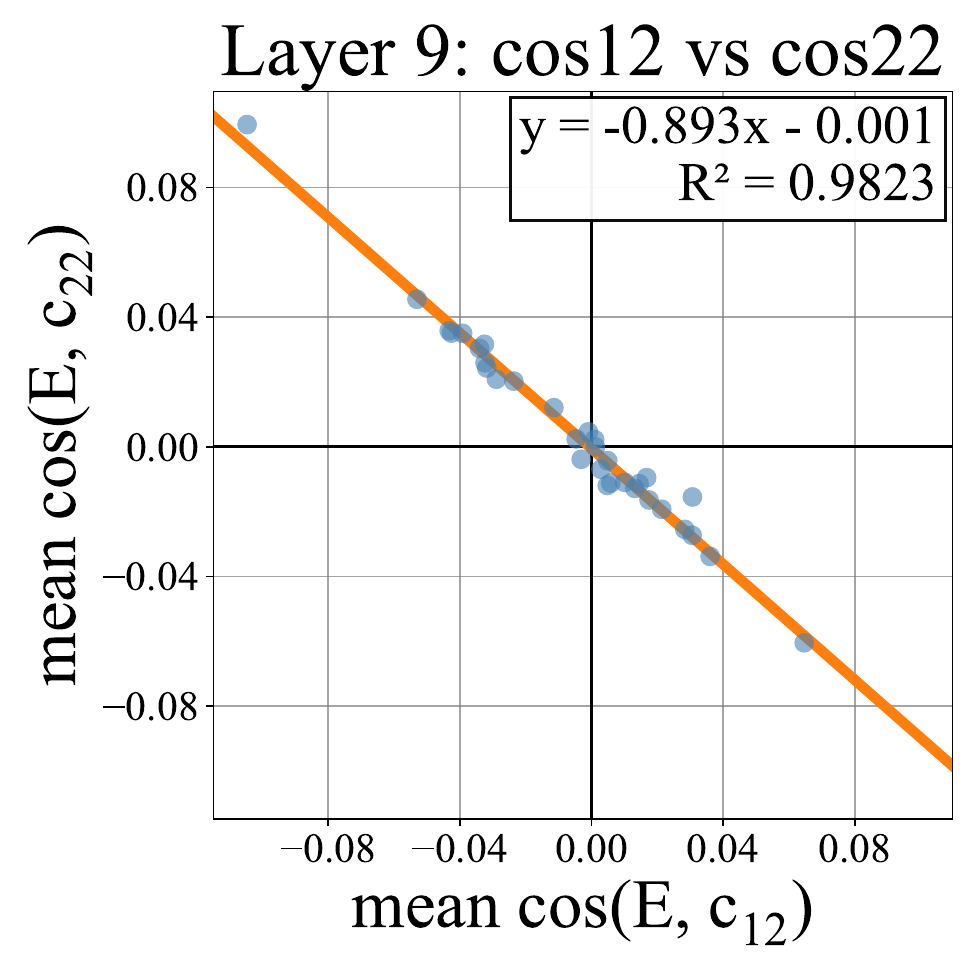}
        \caption{$\cos(\mathbf{E},\mathbf{c}_{12})$ vs. $\cos(\mathbf{E},\mathbf{c}_{22})$}
    \end{subfigure}

    \caption{Head-level mean alignment relationships at Layer~9 of \textit{Mistral-7B-Instruct-v0.3} on 2WikiMQA. Each point corresponds to one attention head, positioned by its global mean cosine alignment.}
    \label{M/Layer9}
\end{figure}

\begin{figure}
    \centering
    \begin{subfigure}{0.32\linewidth}
        \centering
        \includegraphics[width=\linewidth]{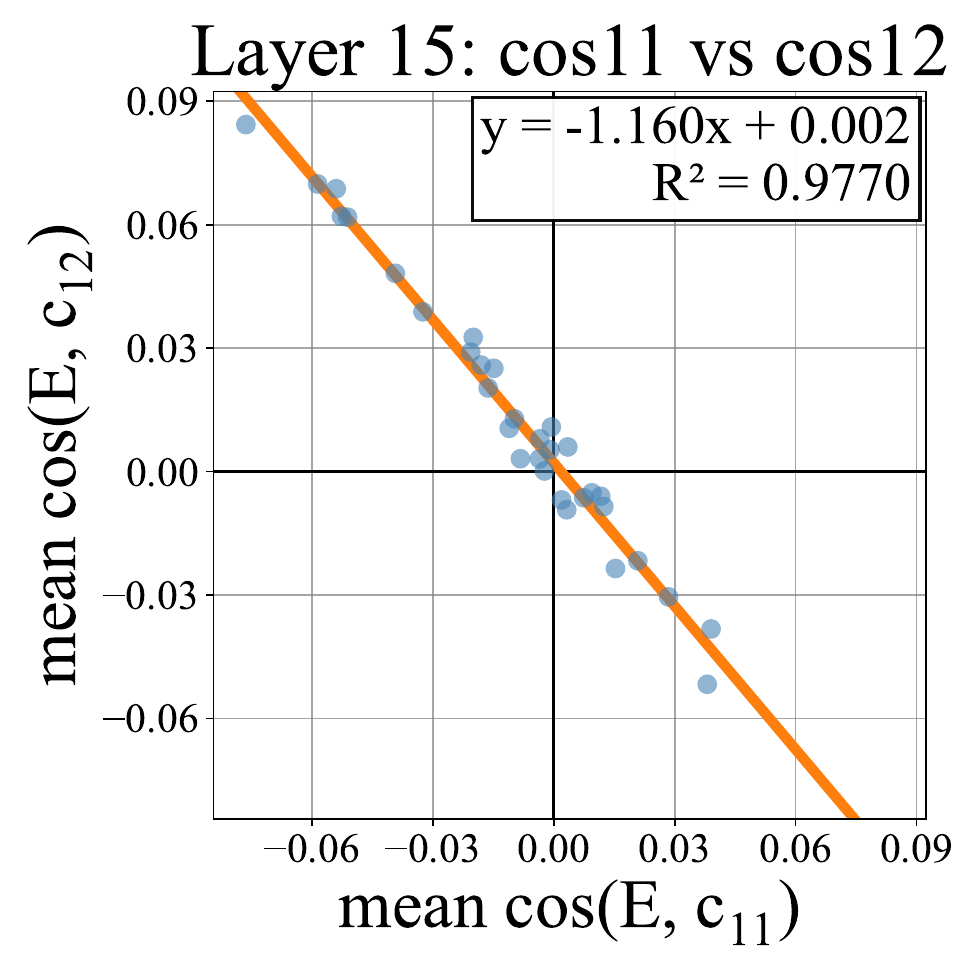}
        \caption{$\cos(\mathbf{E},\mathbf{c}_{11})$ vs. $\cos(\mathbf{E},\mathbf{c}_{12})$}
    \end{subfigure}\hfill
    \begin{subfigure}{0.32\linewidth}
        \centering
        \includegraphics[width=\linewidth]{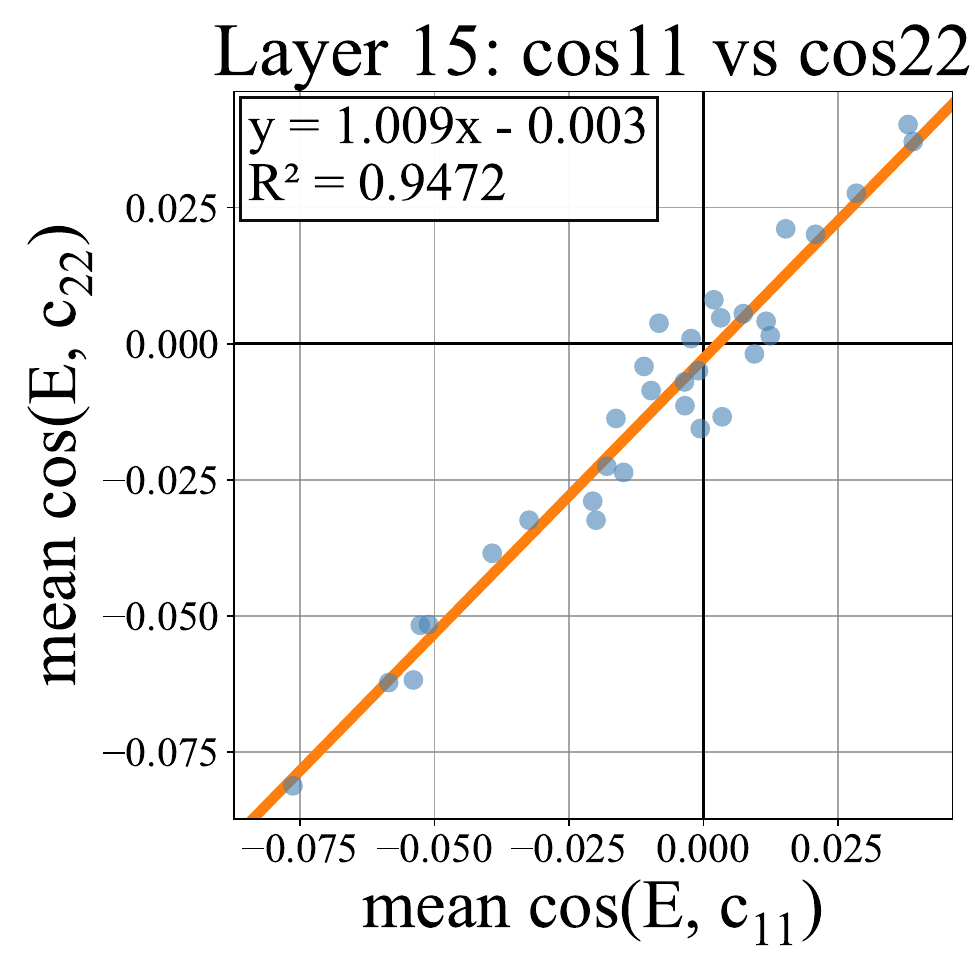}
        \caption{$\cos(\mathbf{E},\mathbf{c}_{11})$ vs. $\cos(\mathbf{E},\mathbf{c}_{22})$}
    \end{subfigure}\hfill
    \begin{subfigure}{0.32\linewidth}
        \centering
        \includegraphics[width=\linewidth]{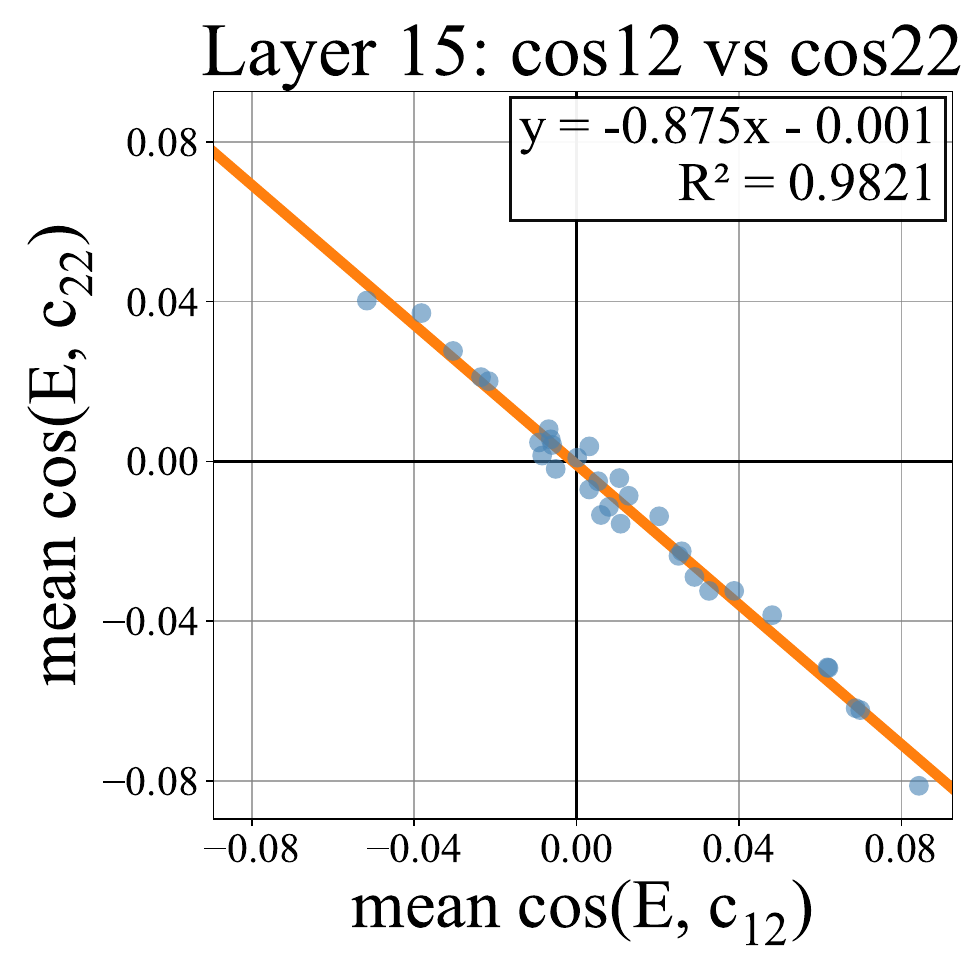}
        \caption{$\cos(\mathbf{E},\mathbf{c}_{12})$ vs. $\cos(\mathbf{E},\mathbf{c}_{22})$}
    \end{subfigure}

    \caption{Head-level mean alignment relationships at Layer~15 of \textit{Mistral-7B-Instruct-v0.3} on 2WikiMQA.
Each point corresponds to one attention head, positioned by its global mean cosine alignment.}
    \label{M/Layer15}
\end{figure}

\begin{figure}
    \centering
    \begin{subfigure}{0.32\linewidth}
        \centering
        \includegraphics[width=\linewidth]{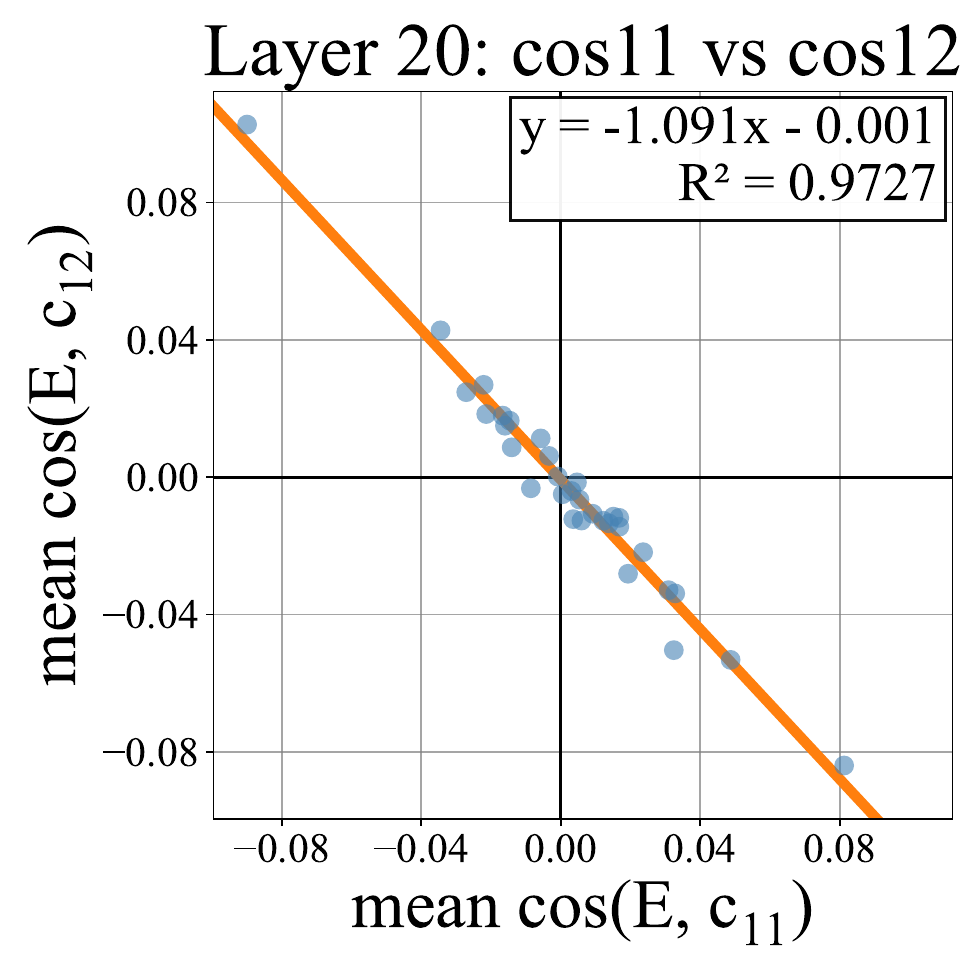}
        \caption{$\cos(\mathbf{E},\mathbf{c}_{11})$ vs. $\cos(\mathbf{E},\mathbf{c}_{12})$}
    \end{subfigure}\hfill
    \begin{subfigure}{0.32\linewidth}
        \centering
        \includegraphics[width=\linewidth]{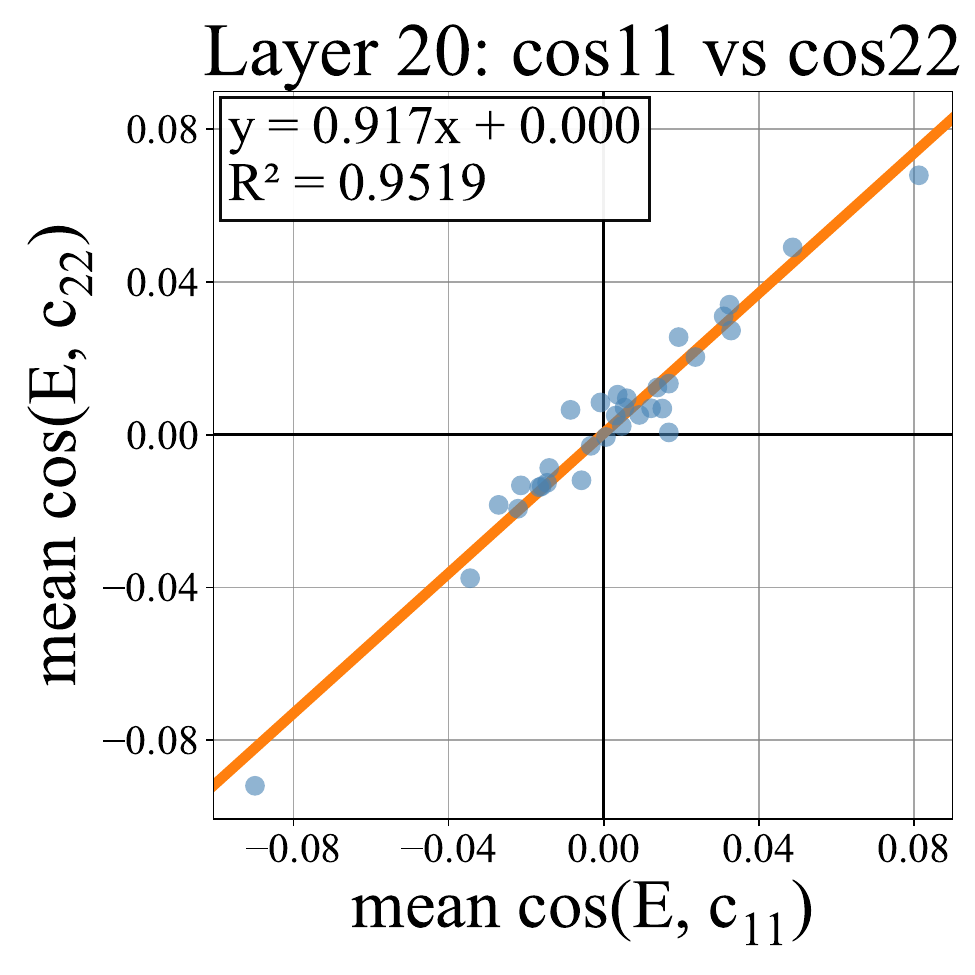}
        \caption{$\cos(\mathbf{E},\mathbf{c}_{11})$ vs. $\cos(\mathbf{E},\mathbf{c}_{22})$}
    \end{subfigure}\hfill
    \begin{subfigure}{0.32\linewidth}
        \centering
        \includegraphics[width=\linewidth]{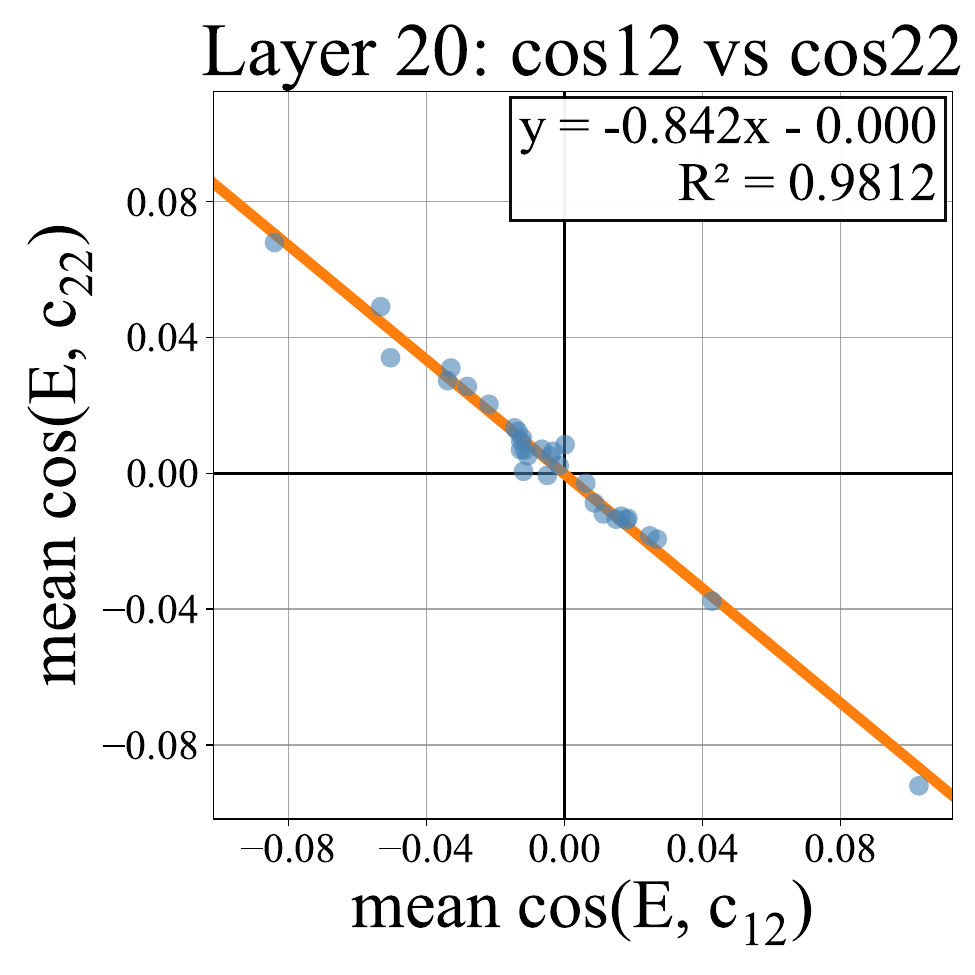}
        \caption{$\cos(\mathbf{E},\mathbf{c}_{12})$ vs. $\cos(\mathbf{E},\mathbf{c}_{22})$}
    \end{subfigure}

    \caption{Head-level mean alignment relationships at Layer~20 of \textit{Mistral-7B-Instruct-v0.3} on 2WikiMQA.
Each point corresponds to one attention head, positioned by its global mean cosine alignment.}
    \label{M/Layer20}
\end{figure}

In Fig.~\ref{llama/Layer5}, Fig.~\ref{llama/Layer22}, Fig.~\ref{M/Layer9}, Fig.~\ref{M/Layer15}, Fig.~\ref{M/Layer20}, we provide more head-level alignment illustrations.

\section{Theoretical Analysis of Eq.~\ref{eq:cos_sign_relation_supp1}}
\label{C}

For a fixed attention head and a query position $q$, the head output is
\begin{equation}
\mathbf o=\sum_i \alpha_i \mathbf v_i,
\qquad
\alpha_i=\mathrm{softmax}(z_i),
\qquad
z_i=\tfrac{1}{\sqrt{d_k}}\,\mathbf q^\top \mathbf k_i .
\label{eq:attn_def_app}
\end{equation}
Define the gradient of the loss with respect to the head output as
\begin{equation}
\mathbf E \triangleq \frac{\partial L}{\partial \mathbf o}.
\label{eq:E_def_app}
\end{equation}
During training, the model is optimized by minimizing the empirical risk
\begin{equation}
\min_{\theta}\ \frac{1}{N}\sum_{n=1}^{N} L^{(n)}(\theta).
\label{eq:L}
\end{equation}
Importantly, $\mathbf E$ is not generated in response to any particular local residual $\mathbf v_i-\mathbf o$.
Rather, it is the head-level first-order descent signal induced by empirical risk minimization.
Concretely, $\mathbf E$ is determined by the downstream network and the final objective through backpropagation;
in a statistical sense, it aggregates gradient information across many training samples, positions, and contexts,
instead of reflecting an instantaneous reaction to a single token or a single value residual.

\paragraph{Head-external vs. head-internal decomposition of key gradients.}
By the chain rule, the gradient with respect to a key $\mathbf k_i$ is
\begin{align}
\frac{\partial L}{\partial \mathbf k_i}
&=
\underbrace{\frac{\partial L}{\partial \mathbf o}}_{\text{head-external } \mathbf E^\top}
\underbrace{\frac{\partial \mathbf o}{\partial \alpha_i}
\frac{\partial \alpha_i}{\partial z_i}
\frac{\partial z_i}{\partial \mathbf k_i}}_{\text{head-internal}}
\nonumber\\
&=
\underbrace{\mathbf E^\top}_{\text{head-external}}
\Bigl(\underbrace{\mathbf v_i-\sum_j \alpha_j \mathbf v_j}_{\text{head-internal residual }(\mathbf v_i-\mathbf o)}\Bigr)\,
\underbrace{\alpha_i(1-\alpha_i)}_{\text{head-internal}}\,
\underbrace{\frac{1}{\sqrt{d_k}}\,\mathbf q}_{\text{head-internal common direction}}
\nonumber\\
&=
\frac{\alpha_i(1-\alpha_i)}{\sqrt{d_k}}\,
\Bigl(\underbrace{\mathbf E^\top(\mathbf v_i-\mathbf o)}_{\text{head-external}\times\text{head-internal: scalar projection}}\Bigr)\,
\underbrace{\mathbf q}_{\text{head-internal common direction}} .
\label{eq:key_grad_full}
\end{align}

Eq.~\ref{eq:key_grad_full} provides a clear decomposition between \emph{head-external} and \emph{head-internal} factors.
Here $\mathbf E$ is the head-external gradient propagated from downstream modules and the final loss;
it specifies the first-order descent direction of the head output $\mathbf o$ under the global training objective,
and does not correspond one-to-one to any specific local residual.
In contrast, $(\mathbf v_i-\mathbf o)$ is the head-internal residual induced by the current context of this head,
and different positions only modulate the update through the scalar projection
$s_i=\mathbf E^\top(\mathbf v_i-\mathbf o)$.

More importantly, Eq.~\ref{eq:key_grad_full} shows that within the same head, the gradients of all keys $\mathbf k_i$
are always colinear with $\mathbf q$; positional differences appear only in the magnitude
$\alpha_i(1-\alpha_i)\,s_i$.
Near convergence, as the overall gradient magnitude decreases, these local modulation terms become mild:
not only is the scale $|s_i|$ suppressed, but the angular variation between $\mathbf E$ and the heterogeneous residuals
$(\mathbf v_i-\mathbf o)$ also becomes more stable.
This aligns with our motivation that $(\mathbf v_i-\mathbf o)$ exhibits intrinsic heterogeneity, while $\mathbf E$ acts as a shared,
approximately unbiased descent direction that does not favor any specific token residual.
Equivalently, $\mathbf E$ tends to induce an approximately consistent projection relationship onto the residual subspace
across contexts.

\paragraph{(i) $\cos(\mathbf E,\mathbf {c}_{11}) \approx \cos(\mathbf E,\mathbf {c}_{22})$.}
Consider the empirical risk objective in Eq.~\ref{eq:L}.
Around a converged solution, the backpropagated signal $\mathbf E$ primarily reflects the head output's first-order descent
direction under the global objective, rather than amplifying a particular local position.
From Eq.~\ref{eq:key_grad_full}, within a head, all key updates share the same head-external factor $\mathbf E$ and the common direction $\mathbf q$,
while positional differences enter only through $s_i=\mathbf E^\top(\mathbf v_i-\mathbf o)$.
As the overall gradient magnitude shrinks near convergence, the variation of these projections across neighboring positions is simultaneously reduced,
so the angular relation (i.e., the angle with $\mathbf E$) becomes comparable.
Thus, while $s_m$ and $s_{m+1}$ need not be exactly equal, their directional behavior is similar in the sense of cosine alignment.

Next we examine the directional properties of the second-order coefficients. By definition,
\begin{equation}
\mathbf c_{11}=\alpha_m(1-2\alpha_m)(\mathbf v_m-\mathbf o),\qquad
\mathbf c_{22}=\alpha_{m+1}(1-2\alpha_{m+1})(\mathbf v_{m+1}-\mathbf o),
\end{equation}
so $\mathbf c_{11}$ and $\mathbf c_{22}$ are exactly parallel to their corresponding local residuals
$\mathbf v_m-\mathbf o$ and $\mathbf v_{m+1}-\mathbf o$, differing only by scalar factors.
In our merging strategy, we only merge adjacent keys with small accumulated attention mass, where typically $\alpha_i<\tfrac12$,
hence $\alpha_i(1-2\alpha_i)>0$.
This factor is strictly positive and therefore preserves direction, merely rescaling the magnitude.
Combining this with the near-convergence stability that $\mathbf E$ responds with similar angular behavior across different residuals,
we obtain
\begin{equation}
\cos(\mathbf E,\mathbf {c}_{11})
=
\cos(\mathbf E,\mathbf {v}_{m}-\mathbf o),
\qquad
\cos(\mathbf E,\mathbf {c}_{22})
=
\cos(\mathbf E,\mathbf {v}_{m+1}-\mathbf o),
\end{equation}
and consequently,
\begin{equation}
\cos(\mathbf E,\mathbf {c}_{11}) \approx \cos(\mathbf E,\mathbf {c}_{22}).
\end{equation}

\paragraph{Sign relation with the off-diagonal softmax Hessian.}
The softmax Hessian has a fixed sign structure: off-diagonal entries are always negative,
\begin{equation}
\frac{\partial^2 \alpha_i}{\partial z_j\,\partial z_k}
=
\begin{cases}
\alpha_i(1-2\alpha_i), & i=j=k,\\[2pt]
-\alpha_i\alpha_j, & i\neq j,
\end{cases}
\label{eq:softmax_hessian_sign}
\end{equation}
so the adjacent coupling term in the value space can be written as
\begin{equation}
\mathbf c_{12}
=
-\alpha_m\alpha_{m+1}\bigl[(\mathbf v_m-\mathbf o)+(\mathbf v_{m+1}-\mathbf o)\bigr].
\label{eq:c12_def_app}
\end{equation}
Let $\mathbf r_m\triangleq \mathbf v_m-\mathbf o$ and $\mathbf r_{m+1}\triangleq \mathbf v_{m+1}-\mathbf o$. Then
\begin{equation}
\mathbf c_{11}=a_m\mathbf r_m,\quad
\mathbf c_{22}=a_{m+1}\mathbf r_{m+1},\quad
\mathbf c_{12}=-b(\mathbf r_m+\mathbf r_{m+1}),
\label{eq:cij_r_form_app}
\end{equation}
which shows that $\mathbf c_{12}$ is a (negative) linear combination of the two residual directions and lies in the same local residual subspace in value space.
Putting the above together yields
\begin{equation}
\cos(\mathbf E,\mathbf{c}_{11})
\;\approx\;
\cos(\mathbf E,\mathbf{c}_{22})
\;\approx\;
-\cos(\mathbf E,\mathbf{c}_{12}).
\label{eq:cos_sign_relation_supp}
\end{equation}
This is also consistent with our empirical observations.

\end{document}